\documentclass[lettersize,journal]{IEEEtran}
\usepackage{amsmath,amsfonts}
\usepackage{algorithmic}
\usepackage{algorithm}
\usepackage{array}
\usepackage[caption=false,font=normalsize,labelfont=sf,textfont=sf]{subfig}
\usepackage{textcomp}
\usepackage{stfloats}
\usepackage{url}
\usepackage{verbatim}
\usepackage{graphicx}
\usepackage{cite}
\usepackage{color}
\usepackage{hyperref}
\usepackage{multirow}
\usepackage{pgfplots}
\pgfplotsset{compat=1.18} % Kompatibilitätsmodus

\begin{document}

\title{Machine Learning Models for Soil Parameter Prediction Based on Satellite, Weather, Clay and Yield Data}

\author{Calvin Kammerlander, Viola Kolb, Marinus Luegmair$^1$, Lou Scheermann, Maximilian Schmailzl,\\ Marco Seufert, Jiayun Zhang, Denis Dali\'c$^2$, Torsten Schön$^3$

\vspace{10pt}

$^1$ orcid: 0000-0001-9022-8428\\
$^2$ MI4People, Munich, Germany\\ 
$^3$ Technische Hochschule Ingolstadt, Ingolstadt, Germany, orcid: 0000-0001-5763-3392\\

\vspace{10pt}

\today

}

\markboth{AgroLens Project, Technische Hochschule Ingolstadt, AKI-23-24}%
{}

%\IEEEpubid{0000--0000/00\$00.00~\copyright~2024}

\maketitle

\begin{abstract}

Efficient nutrient management and precise fertilization are essential for advancing modern agriculture, particularly in regions striving to optimize crop yields sustainably. The AgroLens project endeavors to address this challenge by developing Machine Learning (ML)-based methodologies to predict soil nutrient levels without reliance on laboratory tests. By leveraging state of the art techniques, the project lays a foundation for actionable insights to improve agricultural productivity in resource-constrained areas, such as Africa.
The approach begins with the development of a robust European model using the LUCAS Soil dataset and Sentinel-2 satellite imagery to estimate key soil properties, including phosphorus, potassium, nitrogen, and pH levels. This model is then enhanced by integrating supplementary features, such as weather data, harvest rates, and Clay AI-generated embeddings.
This report details the methodological framework, data preprocessing strategies, and ML pipelines employed in this project. Advanced algorithms, including Random Forests, Extreme Gradient Boosting (XGBoost), and Fully Connected Neural Networks (FCNN), were implemented and fine-tuned for precise nutrient prediction. Results showcase robust model performance, with root mean square error values meeting stringent accuracy thresholds.
By establishing a reproducible and scalable pipeline for soil nutrient prediction, this research paves the way for transformative agricultural applications, including precision fertilization and improved resource allocation in under-resourced regions like Africa.

\end{abstract}

\begin{IEEEkeywords}
Machine Learning, Soil Prediction, Satellite Data, Weather Data, Clay Model, Yield Data, Fertilization Recommendation
\end{IEEEkeywords}

\section[Introduction]{Introduction}
Efficient and sustainable agricultural practices are pivotal in addressing global challenges such as food security, resource scarcity, and environmental degradation. With the increasing demand for higher agricultural productivity, precise fertilization and effective nutrient management have become critical components in modern farming systems \cite{WEIL2017}. 
ML has emerged as a transformative technology in agriculture, providing robust decision-support tools to optimize resource utilization. ML-driven solutions enable the analysis of complex datasets, uncovering insights that traditional methods often overlook. As highlighted by \cite{Ennaji2023}, nutrient management plays a central role in ensuring optimal crop development, requiring precise knowledge of key soil nutrients such as nitrogen, phosphorus, and potassium, and others.

\subsection{Motivation}
Traditional soil testing methods are expensive and inaccessible to many farmers, particularly in under-resourced regions like Africa. By harnessing readily available satellite imagery and advanced ML algorithms, the AgroLens project seeks to democratize soil nutrient estimation, making precision agriculture more accessible and impactful.

\subsection{Objectives}
The overarching goal of the AgroLens project is to develop a scalable and accurate methodology for predicting soil nutrients using non-invasive data sources. This involves:
\begin{itemize}
    \item Constructing ML models capable of estimating the key soil property scores for pH level, Phosphorus, Nitrogen, and Potassium.
    \item Enhancing model performance through the integration of diverse data sources, including local weather data and harvest rates.
\end{itemize}

\subsection{Reasons for the Use of Machine Learning}
ML techniques are well suited to agricultural applications due to their ability to handle large, heterogeneous datasets and uncover complex, non-linear relationships between variables. For instance, ML models can integrate satellite-derived spectral indices, climate variables, and other geospatial data to produce reliable predictions of soil nutrient levels. This capability makes ML a powerful tool for addressing the variability and unpredictability inherent in agricultural systems.

\subsection{Overview of the Approach}
The AgroLens project employs a structured two-phase methodology:
\begin{enumerate}
    \item \textbf{Model Development}: Leveraging European datasets (e.g., LUCAS Soil and Sentinel-2 imagery) to create a baseline model for soil nutrient estimation.
    \item \textbf{Feature Enrichment}: Incorporating supplementary data sources such as weather data, harvest rates, and Clay embeddings to improve model accuracy.
\end{enumerate}
By following this approach, AgroLens aims to establish a reproducible and scalable framework for soil nutrient prediction, laying the groundwork for future work in precision agriculture.\\

\section[Methodology]{Methodology}
In this study, ML is used to predict soil properties using different types of data. The input data includes satellite images, weather data, harvest rates and Clay embeddings, while nutrient scores form the target data. The training pipeline includes data pre-processing, feature selection and model evaluation using Root Mean Square Error (RMSE).

\subsection[Methodology-ML-for-Soil]{Machine Learning for Soil Prediction}
ML is widely applied across numerous domains, including agriculture, where it has shown particular promise for soil prediction—especially in African contexts \cite{REJEB2022, KUDAMA2021}. Various ML techniques, such as Random Forest, XGBoost, FCNN, Convolutional Neural Networks, and Support Vector Machines, have been explored for nutrient management in soils. Typical input data include soil properties, climatic variables, and satellite imagery, enabling accurate estimations of nutrient levels \cite{Ennaji2023}.

\subsection[Methodology-Data-Types]{Data Types}
This section provides an overview of the data types used in the project, beginning with the input datasets and concluding with the target variables.\\

\subsubsection[Methodology-Input-Data]{Input Data}
\paragraph{Satellite Data}
Satellite imagery serves as a primary source of spatial information in AgroLens, capturing reflectance values across multiple spectral bands. By analyzing these reflectance patterns, one can infer soil conditions, vegetation health, and other environmental variables.

The Sentinel-2 mission was launched in 2015 and includes satellites that revisit a given location every five days. It offers 13 spectral bands with resolutions ranging from 10 to 60 m, covering portions of the visible, near-infrared, and shortwave infrared spectrum. This project uses Sentinel-2 Level~2A images, which undergo atmospheric correction to reduce distortions and provide more accurate surface reflectance values. Using Level~2A data reduces the number of available bands to 12 \cite{Sentinel2}. Sentinel-2 is chosen for its high revisit frequency and data quality, improving the chances of obtaining cloud-free images aligned with the soil-sample collection dates.

Landsat~7 and Landsat~8 were launched in 1999 and 2013, respectively, each following a 16-day repeat cycle. This longer interval limits the frequency of observations compared to Sentinel-2, making time-sensitive analyses more challenging. Landsat~7 provides eight spectral bands with spatial resolutions between 15 and 60 m, whereas Landsat~8 adds three more bands (for a total of 11). However, the two thermal infrared bands of Landsat~8 are excluded here to match the Sentinel-2 band set, leaving nine bands with 15--30~meter resolutions. Both Landsat~7 and Landsat~8 also offer Level~2 atmospheric-corrected products. \cite{Landsat7,Landsat8}

\paragraph{Neighboring Pixels}
In addition to the central pixel patch, neighboring pixels around each sampled location are considered to capture more spatial context. The premise is that pixel-to-pixel variations in reflectance might reveal subtle gradients related to soil properties, vegetation cover, or microtopography.

\paragraph{BigEarthNet v2.0}
BigEarthNet v2.0 is a benchmark dataset derived from Sentinel-1 and Sentinel-2 satellite imagery \cite{BigEarthNetDatasetWebsite2025}, published in conjunction with the work of Clasen et al. \cite{BigEarthPaper2024}. It is designed for Earth observation, focusing on land-cover classification and environmental monitoring. The dataset contains annotated image pairs from several European countries, encompassing diverse land-use and land-cover classes (e.g., agricultural fields, forests, urban areas, water bodies) \cite{BigEarthPaper2024}.
Additionally, the BigEarthNet initiative provides pre-trained models \cite{BigEarthNetModelsWebsite2025}, which can be used to classify land-use and land-cover types in regions beyond those originally included in BigEarthNet v2.0. However, this paper concentrates on predicting soil nutrients rather than classifying land use. Because the training data exclusively represent agricultural areas, land-use classification was deemed unnecessary. Nonetheless, incorporating such classification could be a useful extension in future enhancements of this system, ensuring that predictions are only generated for valid agricultural zones.

\paragraph{Weather Data}
Aligning weather data to the same temporal window as satellite imagery ensures that the satellite reflectance can be interpreted in the context of concurrent weather conditions.

\paragraph{Harvest Rates}
The Food and Agriculture Organization (FAO) publishes comprehensive yield statistics that can be leveraged to approximate crop productivity and, indirectly, soil fertility indicators across different regions. These datasets generally include average yield estimates for major crops (e.g., wheat, maize, rice) at subnational or national levels.
Harvest rates can serve as a proxy for soil fertility when more direct soil measurements are unavailable or incomplete. Highly productive regions typically correlate with better soil conditions and nutrient availability. Additionally, integrating harvest rates can help contextualize local satellite pixel values, as known crop outputs may offer insights into whether observed reflectance patterns align with high- or low-fertility conditions.

\paragraph{Data Inspection}
Several measures are taken during data preprocessing to ensure input quality and consistency. Rather than discarding rows with missing values, only potassium measurements that fall below the limit of detection are imputed (see Section \ref{A-Lucas-Soil}).
Potential outliers are identified by generating histograms for each feature. Although a small fraction (less than 0.1\% of the total dataset) lies noticeably outside the main distribution, these points are retained under the assumption that a robust model can handle such rare extremes.
Lastly, the Pearson product-moment correlation coefficient is computed,
\[
R_{ij} = \frac{C_{ij}}{\sqrt{C_{ii}C_{jj}}},
\]
where \(C\) is the covariance matrix. Correlation values approaching 1.0 suggest strong linear relationships, while those near 0.0 indicate minimal or no linear association. This analysis assists in detecting redundant features and gauging overall data quality.\\

\subsubsection[Methodology-Target-Data]{Target Data} \label{ChapterMethTargetData}
As part of the AgroLens project, the primary soil properties identified for initial prediction are pH levels, phosphorus, potassium, and nitrogen. An evaluation of soil data sources is carried out following \cite{Hengl2021}, which mainly relies on AfSIS, and is extended to include other soil data sources beyond Africa, such as LUCAS \cite{LUCAS2018} and WoSIS \cite{WoSIS2024}. The resulting findings, along with assessments of their suitability for mapping against satellite data, appear in Table~\ref{tab_soil_data}.
A key requirement for mapping soil data to satellite imagery is the availability of timestamps for the soil sampling event that are sufficiently accurate (within approximately one month), as well as overlap with the relevant satellite-data coverage. Although this threshold may seem self-evident, it poses a significant challenge in the case of many African soil samples.
\begin{table}[htbp]
\caption{Soil data sets with timestamp attribute and Landsat 8 and Sentinel-2 coverage}
\begin{center}
\begin{tabular}{|c@{}|c@{}|c@{}|c@{}|c@{}|c@{}|}
\hline
\multirow{2}{*}{\textbf{Data set}}      & \multirow{2}{*}{\textbf{\begin{tabular}[c]{@{}c@{}}Time-\\ frame\end{tabular}}} & \multirow{2}{*}{\textbf{\begin{tabular}[c]{@{}c@{}}Time-\\ stamp\end{tabular}}} & \multirow{2}{*}{\textbf{Profiles}} & \textbf{Landsat 7/8}  & \textbf{Sentinel-2}   \\
                                        &                                                                                 &                                                                                 &                                    & \textbf{$(>03/2008)$} & \textbf{$(>06/2015)$} \\ \hline
LUCAS \cite{LUCAS2018} & 2018                                                                            & Yes                                                                             & 18,984                             & 18,471                & 18,471                \\
AfSIS \cite{AfSIS2020} & 2009-2018                                                                       & No                                                                              & 20,704                             & 0                     & 0                     \\
WoSIS \cite{WoSIS2024} & 1920-2023                                                                       & Partially                                                                       & 228,000                            & 3,641                 & 0                     \\ \hline
\end{tabular}
\label{tab_soil_data}
\end{center}
\end{table}

\subsection[Methodology-ML-Models]{Machine Learning Models}
The prediction of soil nutrient levels using satellite data combined with additional information, such as weather data, represents a typical regression problem. Consequently, three model variants are selected for each nutrient type to utilize the specific strengths of each model in predicting the corresponding soil nutrient level and to perform a comparative performance analysis.\\

\subsubsection[Methodology-XGBoost]{XGBoost}
XGBoost is a tree boosting system, published by \cite{XGBoost2016}. XGBoost builds trees sequentially, with each new tree aiming to correct the errors of the previous tree. The final prediction is obtained by combining all trees. The method is known for its high accuracy and computational efficiency. Furthermore. it is robust to large and noisy datasets. The built-in regularization techniques help mitigate overfitting.\\

\subsubsection[Methodology-Fully-Connected-Neuronal-Network]{Fully Connected Neuronal Network}
A FCNN is a deep learning architecture designed to model non-linear interactions among input features \cite{Goodfellow-et-al-2016}. By fully connecting every neuron in each layer to all neurons in adjacent layers, FCNNs can capture a wide range of multivariate relationships between soil properties and environmental factors, making them particularly well-suited for tasks requiring complex pattern discovery.\\

\subsubsection[Methodology-Random-Forest]{Random Forest}
Random Forest is an ensemble method comprising multiple decision trees, initially introduced by \cite{RandomForest2001}. It is recognized for its robustness and interpretability, offering reliable predictions while maintaining relatively low computational demands compared to more complex models. This approach is especially suitable for scenarios with limited training data.

\subsection[Methodology-Training-Pipeline]{Training Pipeline}
The machine learning models are run on a server equipped with 8~CPU cores, 32~GB of RAM, and an NVIDIA RTX~3060 GPU with 6~GB of VRAM. Remote collaboration is facilitated through isolated Docker containers that provide GPU access, ensuring an efficient and organized workflow. Preprocessed datasets and resulting models are stored and versioned on the same server, while all code is published to a Git repository for collaboration and version control (\href{https://github.com/cvims/AgroLens}{https://github.com/cvims/AgroLens}).\\

\subsubsection[Methodology-Data-Preprocessing]{Data Preprocessing}
In this section the mandatory data preprocessing is described.
\paragraph{Data Collection}
The training, test, and validation data for this project is collected from multiple sources. 
LUCAS~2018 TOPSOIL data is obtained by submitting an official request form to the European Soil Data Centre (ESDAC), which then provides the dataset as structured CSV tables \cite{LUCAS2018}. This data is further preprocessed through Python scripts. Satellite imagery from Sentinel-2 and Landsat~8 is acquired via the Copernicus Data Space API through automated requests spanning several days. Additionally, weather and climate information is sourced from the OpenWeatherMap API, which requires a paid subscription for full access to its datasets. These diverse data sources are carefully integrated to support the project objectives.
Weather data is retrieved using the OpenWeather API’s Time Machine endpoint, which delivers historical weather information for specified geographic locations and dates. For each row in the dataset, latitude, longitude, and the date are extracted and used to construct API queries. A redundancy check ensures that API calls are only made for rows lacking weather data, thus minimizing costs. The returned data is parsed to extract key metrics such as temperature, humidity, wind speed, and sunrise/sunset times, which are then appended to the dataset. For details, see \href{www.openweathermap.org/}{www.openweathermap.org/}.
Harvest rates are downloaded from the FAO GAEZ portal by selecting the “Theme~5: Actual Yields and Production” dataset and obtaining the relevant GeoTIFF files. These georeferenced files are then opened in a GIS environment to extract pixel values for each set of latitude and longitude coordinates. In this manner, yield scores become available for spatially explicit analysis of agricultural productivity. Refer to \href{https://gaez.fao.org/}{https://gaez.fao.org/} for more information.
Finally, WoSIS 2023 snapshot data is openly available as a zipped dataset from the web, offering structured CSV and TSV tables that are further processed through Python scripts \cite{WoSIS2023_snapshot}.

\paragraph{Normalization}
To ensure consistency and improve the performance of ML models, data normalization is implemented as part of the preprocessing pipeline. The AgroLens project adopts a min-max normalization strategy, which scales the input features to a range of [0, 1]. This method preserves the relationships between values while standardizing the dataset for computational efficiency.
The normalization process is designed to retain the integrity of the original data by maintaining separate tables for raw and normalized values:
\begin{itemize}
    \item \textbf{Original Data Table}: Contains the unmodified, raw data as collected from various sources.
    \item \textbf{Normalized Data Table}: A transformed version of the original data, where each feature value \(x\) is scaled using the formula:
    \[
    x_{\text{normalized}} = \frac{x - x_{\text{min}}}{x_{\text{max}} - x_{\text{min}}}
    \]
    Here, \(x_{\text{min}}\) and \(x_{\text{max}}\) are the minimum and maximum values of the feature, respectively. 
\end{itemize}
The target features, which represent soil nutrient levels, are not normalized as the goal is to predict these values on their original scale, ensuring meaningful and interpretable outputs.
This approach ensures that the original dataset remains untouched for verification and comparison purposes, while the normalized dataset is utilized for training and validating ML models. The separation also provides flexibility for further analysis and troubleshooting during model development and evaluation.
By applying this normalization technique to the input features, the project mitigates the risk of bias introduced by features with different scales or units, thereby enhancing the stability and accuracy of the predictive models.

\paragraph{Spatial Cross-Validation}\label{ChapterMethodeSCV}
Spatial Cross-Validation (CV) is an approach for evaluating model performance in projects involving spatial data. It is used to take spatial autocorrelation of the dataset into account and mitigates overestimation of model accuracy \cite{SpatialCV_2016}.\newline
Traditional validation methods, which randomly split data into sets for training and testing, often result in spatially proximate samples appearing in both sets. This can result in overly optimistic performance metrics, as similar observations in both sets can artificially inflate model accuracy \cite{SpatialCV_2016}.
Therefore, it is recommended to account for spatial dependence when validating a model using spatial data. There are various approaches for spatial CV. One such method is Grid-based spatial CV. In this approach the dataset is divided into separate, spatial grid cells. The grids can be divided into training, validation and test grids \cite{SpatialCV_2023}.
The choice of the grid cell size is crucial. Smaller grids may result in test datasets sharing similar characteristics with the training datasets, while larger blocks increase the risk that test data is not spatially similar, which could lead to better validation scores for the model \cite{SpatialCV_2016}.\\

\subsubsection{Used ML Tools}\label{Methodology-Used-ML-Tools}
Open-source software is employed throughout this research, except where commercial products are explicitly mentioned. Most computations are performed in Python \cite{Python2025}, a widely recognized and state-of-the-art language for ML. The following Python libraries are also utilized:

\begin{itemize}
    \item \textbf{numpy}—“The fundamental package for scientific computing with Python” \cite{Numpy2025,Harris2020}. 
    \item \textbf{scipy}—Offers advanced mathematical functions for scientific computing \cite{Virtanen2020,Scipy2025}.
    \item \textbf{optuna}—Facilitates efficient hyperparameter tuning, supporting methods like grid search, random sampling, genetic algorithms, Covariance Matrix Adaptation Evolution Strategy (CMA-ES) \cite{Hamano2022}, Gaussian Processes, and Tree-structured Parzen Estimator (TPE) \cite{Bergstra2011}. A dashboard feature provides real-time visualization of optimization progress and parameter sensitivity \cite{0ptuna2019,0ptuna2025,Plotly2025}.
    \item \textbf{pandas}—Handles data-frame preparation and manipulation \cite{Mckinney2010,pandas2020,Pandas2025}.
    \item \textbf{pytorch}—Provides a flexible, high-performance framework for deep learning tasks, used here for the FCNN \cite{Pytorch2025}.
    \item \textbf{scikit-learn}—A widely adopted library offering an assortment of common machine learning algorithms \cite{Pedregosa2011,Scikit2025}.
    \item \textbf{xgboost}—Implements a distributed gradient boosting framework for efficient model training \cite{XGBoost2016,XGBoost2025}.
    \item \textbf{Additional Tools}—Includes \texttt{boto3}, \texttt{gdal}, \texttt{geopandas}, \texttt{joblib}, \texttt{jupyter}, \texttt{OpenCV}, \texttt{rasterio}, \texttt{requests}, and \texttt{shapely}, among others, for specific data processing needs.\\
\end{itemize}

\subsubsection[Methodology-Visualization-Tools]{Visualization Tools}
Visualization tasks are handled by the following Python libraries:

\begin{itemize}
    \item \textbf{matplotlib}—Provides fundamental plotting capabilities for charts, graphs, and figures \cite{Matplotlib2024}.
    \item \textbf{seaborn}—Extends matplotlib with advanced statistical visualizations, such as heat maps and correlation matrices \cite{Waskom2021,Seaborn2025}.\\
\end{itemize}

\section[A]{Model for Europe}
\label{modeleurope}
\begin{figure*}[ht]
    \centering
    \includegraphics[width=\textwidth]{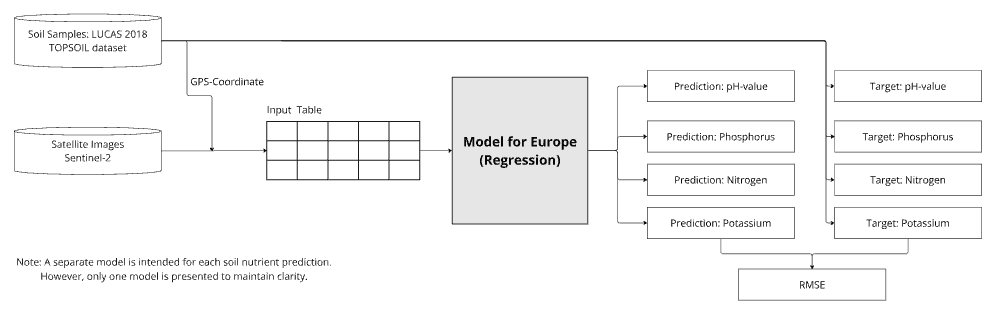}
    \caption{Schematic Concept of the Training Process for Model and Data Usage for the Europe Model}
    \label{fig:03ConceptModelEurope}
\end{figure*}

This section presents a predictive modeling approach that uses satellite images and field-collected soil measurements to estimate key soil parameters. The model integrates Sentinel-2 satellite imagery as input data and soil property measurements from the LUCAS 2018 TOPSOIL dataset as target data. The concept of the model for Europe is visualized in figure \ref{fig:03ConceptModelEurope}.\newline
To ensure robust predictions, three ML models, including XGBoost, FCNN and Random Forest, are implemented and evaluated. Furthermore, different data splitting techniques, such as single split and spatial cross-validation, are explored to assess the impact of spatial dependencies on model performance. The following sections detail the data preprocessing, model selection, hyperparameter optimization, and performance analysis. 

\subsection[A-Data-Set]{Data Set}
The Model for Europe requires input data and target data as detailed in the following section. Additionally, strategies for splitting the dataset are provided.\\

\subsubsection[A-Input-Data]{Input Data}
\paragraph[A-Sentinel]{Sentinel-2}
Sentinel-2 image data for each soil sample is obtained through multiple steps. First, all available satellite datasets matching each soil sample’s GPS coordinates within a 29-day window (14 days before and 14 days after the sample date) are identified by the Copernicus API. Next, only the cloud probability masks for these datasets are downloaded to assess the pixel-level cloud probability within a small radius of the target coordinate. Once the nearest cloud-free dataset is identified, the corresponding 100km × 100km tile is downloaded as a ZIP file, then extracted and placed into a fixed directory structure. Each single-band grayscale image is cropped to 101 × 101 pixels, ensuring that the center pixel aligns precisely with the soil sample’s location. Finally, the dataset folder is stored in the input data directory on the server, systematically organized by date and by GPS latitude and longitude.
This procedure is repeated for every LUCAS SOIL dataset sample, resulting in the download and processing of approximately 20TB of raw image data and thus requiring optimization. Three distinct Python libraries are benchmarked on Sentinel-2 datasets to determine the most efficient method for cropping the images. In this case, OpenCV outperforms both Pillow and scikit-image, completing the task nearly twice as quickly. To reduce resource bottlenecks, a multi-threaded approach is implemented, with each thread dedicated to a specific task (e.g., image cropping, cloud detection, API calls, or data downloads). Temporary files are stored on RAM-disks to improve performance and minimize wear on the servers’ SSDs. With the combined resources of three available servers, the process takes multiple days to complete and ultimately produces approximately 120GB of processed Sentinel-2 images. Afterwards, depending on the specific project model, the required pixels are extracted and saved into a data table using a separate command line script.

\begin{figure}[ht]
    \centering
    \includegraphics[width=1.0\linewidth]{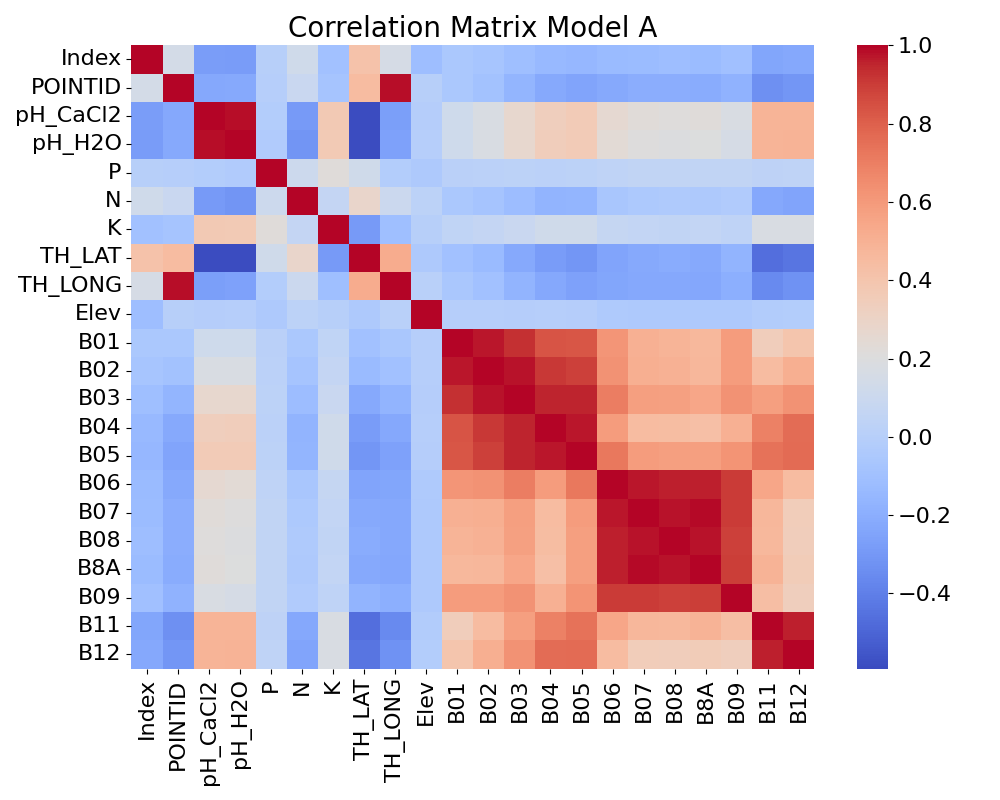}
    \caption{Correlation coefficient matrix for the input data from Sentinel of the Europe model}
    \label{fig:03-CorrCoeff-Input}
\end{figure}

\paragraph{Correlation of the Input Data}\label{Europe-Corr-Input}
In Figure \ref{fig:03-CorrCoeff-Input}, the correlation coefficient matrix for the input data is presented. A pronounced linear correlation is observed among Sentinel-2 bands 1–5, 6–9, and 11–12. This is likely due to all bands scanning the same spatial location at different wavelengths. Despite these high correlations, none of the Sentinel-2 bands are removed to preserve the maximum amount of information, given the relatively small size of the training dataset.
Additionally, a strong correlation is evident between the POINTID and the longitudinal position of the measurement point, presumably because the longitudinal value was used to index the dataset. Since Index and POINTID are not used as training variables, this correlation does not influence the model’s training process.\\

\subsubsection[A-Target-Data]{Target Data}
The Model for Europe target dataset and its information is detailed below.

\begin{figure}[ht]
    \centering
    \includegraphics[width=1\linewidth]{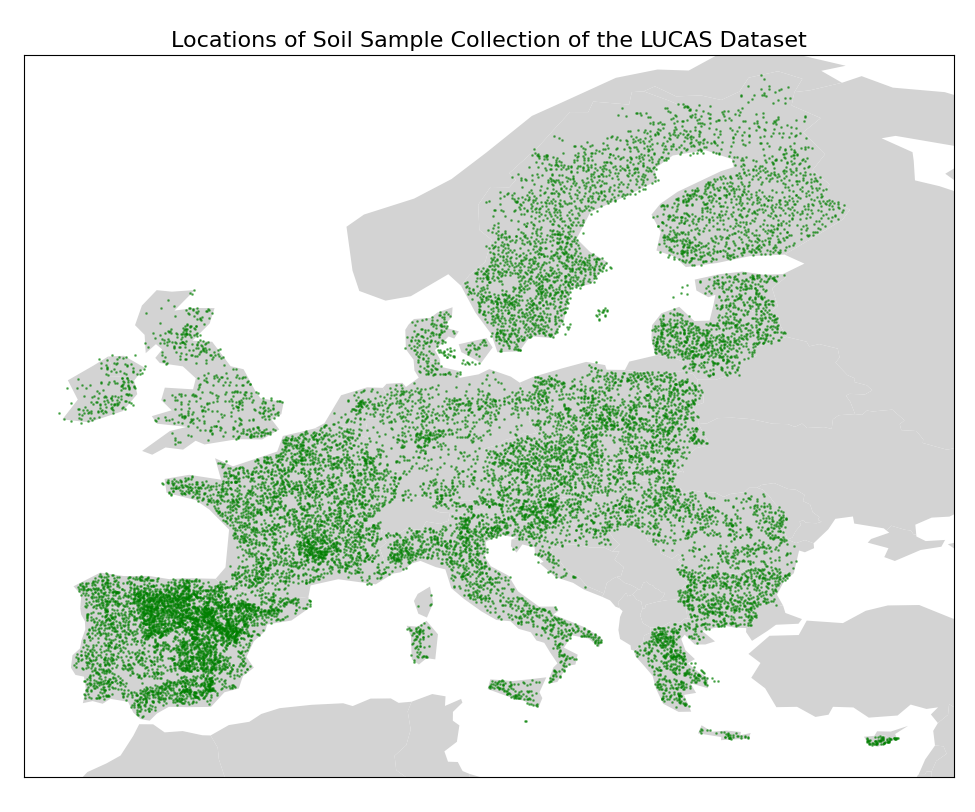}
    \caption{Locations of Soil Sample Collection of the LUCAS 2018 TOPSOIL Dataset}
    \label{fig:LUCASSOilSample}
\end{figure}

\begin{figure}[ht]
    \centering
    \includegraphics[width=1.0\linewidth]{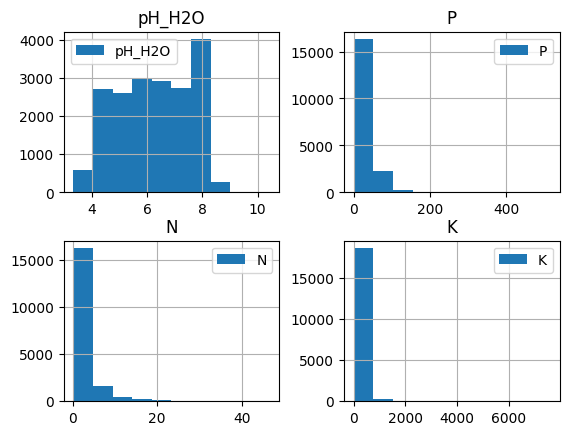}
    \caption{Histogram for the target data nutrients pH, nitrogen, phosphorus and potassium in the LUCAS 2018 TOPSOIL dataset}
    \label{fig:elements-histogram}
\end{figure}

\paragraph{LUCAS 2018 TOPSOIL Dataset}\label{A-Lucas-Soil}
The LUCAS Programme is an area-frame statistical survey organized and managed by Eurostat (the Statistical Office of the European Union) to monitor changes in land use (LU) and land cover (LC) over time across the EU \cite{ESDAC2022}. Since 2006, Eurostat has conducted LUCAS surveys every three years, relying on visual assessments of environmental and structural landscape features at georeferenced control points. These points are located at intersections of a 2km × 2km regular grid spanning the EU, generating approximately one million georeferenced points. In each survey, a subsample of these points is selected for collecting field-based information.
The LUCAS 2018 TOPSOIL dataset comprises detailed soil property measurements from 18,984 samples collected throughout the European Union and the UK. It includes data on pH ($CaCl_2$ and $H_{2}O$), organic carbon, $CaCO_3$, nitrogen, phosphorus, potassium, electrical conductivity (EC), and oxalate-extractable iron and aluminum \cite{LUCAS2018}. Figure \ref{fig:LUCASSOilSample} provides a geographical overview of the data points included in the dataset.
As noted previously, the goal is to predict values for pH ($CaCl_2$ and $H_{2}O$), nitrogen, phosphorus, and potassium. For pH, two measurements are available depending on whether $CaCl_2$ or $H_{2}O$ is used. Each of the four target nutrients has a distinct limit of detection (LOD), defined as the lowest quantity that can be measured with sufficient reliability. Table \ref{table_lucas_soil_lod} summarizes these limits of detection.
\begin{table}[ht]
    \centering
    \caption{Description of the soil fields in the LUCAS 2018 TOPSOIL dataset}
    \begin{tabular}{|l@{}|l@{}|c@{}|c@{}|}
    \hline 
          Field & Description & Unit & LOD\\
    \hline
         pH(CaCl2) & pH measured in a CaCl2 solution & - & 2-10 \\
         pH(H2O) & pH measured in a suspension of soil in water & - & 2-10 \\
         N & Total nitrogen content & g/kg & 0.2 \\
         P & Phosphorus content & mg/kg & 10 \\
         K & Extractable potassium content & mg/kg & 10 \\
    \hline
    \end{tabular}
    \label{table_lucas_soil_lod}
\end{table}
Out of the complete LUCAS 2018 TOPSOIL dataset, 4,945 potassium values fall below the 10mg/kg limit of detection (LOD). These values are imputed by assigning an average estimate between 0 and 10mg/kg, resulting in a constant replacement of 5mg/kg. After imputing potassium, data cleansing and matching with Sentinel-2 imagery reduce the dataset to 18,471 soil samples.
Figure \ref{fig:elements-histogram} shows histograms of the target nutrients. While pH is relatively evenly distributed, nitrogen, phosphorus, and potassium exhibit a pronounced left skew. Along with min–max normalization, log normalization is evaluated for the skewed nutrients (N, P, and K); however, since it does not improve model performance, the min–max normalization approach is ultimately maintained for all four nutrients.

\paragraph{Correlation of the Target Data}
Figure \ref{fig:03-CorrCoeff-Input} also presents the correlation coefficient matrix for the target data. The very high correlation between pH\_H2O and pH\_CaCl2 arises because both values measure pH, albeit via two different methods. This strong correlation reflects the reliability of the pH measurements and suggests that training separate models for each pH variant may not be necessary.
A notable linear correlation between input features and target values only emerges for the pH values and latitude (negative correlation), as well as for Sentinel bands B11 and B12 (positive correlation). These findings indicate that latitude and bands B11 and B12 are likely to be significant predictors of pH.\\

\subsubsection[A-Data Split]{Dataset Split}
The following section describes the procedure for partitioning the dataset, which is used for training and validating the model.

\paragraph[A-Data Split]{Single Split}\label{SectionSingleSplit}
The dataset, containing the previously described input and target data, includes 18,471 samples and is visualized in Figure \ref{fig:LUCASSOilSample}. For training, it is randomly split into an 80:20 ratio, resulting in 14,776 samples in the training set and 3,695 samples in the test set. It is important to note that a single random split does not account for potential spatial dependencies in the data.

\begin{figure}[ht]
    \centering
    \includegraphics[width=1\linewidth]{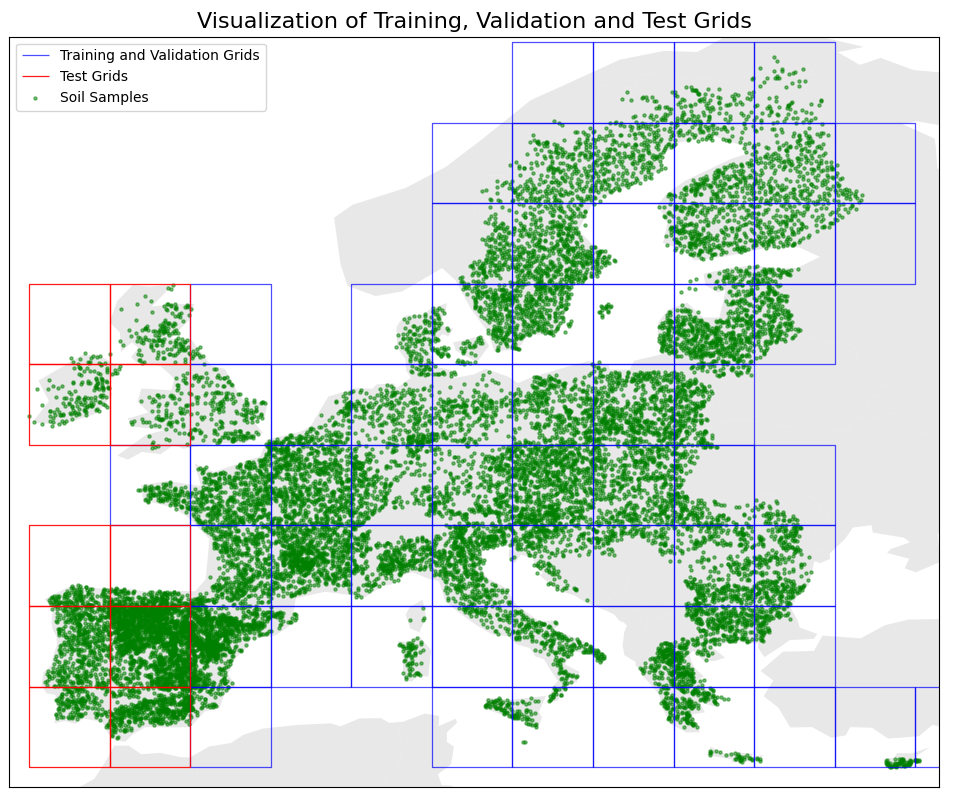}
    \caption{Visualization of Training, Validation and Test Grids (Grid size: 4°× 4°)}
    \label{fig:SCV_Grids}
\end{figure}

\paragraph[A-Data Split]{Spatial Cross Validation}
As described in Section \ref{ChapterMethodeSCV}, a single split does not account for spatial dependencies. Therefore, in addition to the single-split approach, spatial cross-validation (CV) is applied and its effect on performance is evaluated using the XGBoost model. The corresponding results are discussed in Section \ref{Chapter03_ResXgboost}.
To implement spatial CV, data points are grouped into grid cells based on their geographical location. The grid size affects both the total number of grid cells and the number of soil samples within each cell. Multiple grid sizes were considered during this project; this paper focuses on a 4° × 4° grid.
Each grid cell is used to split its contained data into training, validation, and test subsets. The overall goal is an approximate 60:20:20 split of the dataset. Because the grid cells contain different numbers of samples, the ratio can only be approximated. For a 4° × 4° grid, the test dataset includes 3,567 samples, while 15,173 samples remain for training. The training set is then further divided via a 5-fold CV approach, where each fold consists of distinct grid cells.
Based on the 5-fold CV, the RMSE\textsubscript{Average} is computed for the validation data. Additionally, the RMSE\textsubscript{Test} is calculated using the unseen test dataset, which does not appear in any training or validation folds. Figure \ref{fig:SCV_Grids} illustrates how the 4° × 4° grid is partitioned into training, validation, and test cells.

\subsection{Used Model Details}\label{A-Used-Model-Details}
This section provides a detailed model overview for Europe, including hyperparameter optimization for XGBoost, FCNN and Random Forest. It explains which parameters are optimized for each model to achieve the best performance.\\

\subsubsection[A-XGBoost]{XGBoost}
The hyperparameter tuning for the XGBoost model is performed using Optuna, as introduced in Section \ref{Methodology-Used-ML-Tools}. 
The following two properties are pre-defined and not subject to hyperparameter optimization
\begin{itemize}
    \item Evaluation Metric: The RMSE is the selected evaluation metric for model performance.
    \item Tree Construction Method:  The hist algorithm is selected to accelerate the tree construction based on histograms for improved large dataset efficiency.
\end{itemize}
The following hyperparameters are optimized during the hyperparameter tuning process:
\begin{itemize}
    \item Maximum Tree Depth (max\_depth): The value for the maximum depth of the decision trees is optimized in the range from 3 to 12.
    \item Learning Rate (learning\_rate): The learning rate is varied between 0.01 and 0.3 to control the model adaptation rate.
    \item Subsampling Rate (subsample): The fraction of the training data used for model fitting is optimized between 0.6 and 1.0.
    \item Column Sampling (colsample\_bytree): The fraction of features randomly selected for each tree is optimized between 0.6 and 1.0.
    \item Gamma: The regularization term controlling the minimum reduction of the loss for each split is optimized between 0 and 1.
    \item L1 Regularization (reg\_alpha): The L1 regularization parameter is set in the range from 0 to 1 to promote sparser models.
    \item L2 Regularization (reg\_lambda): The L2 regularization parameter is also varied between 0 and 1 to prevent overfitting.
\end{itemize}
An overview of the selected hyperparameters for each XGBoost model is shown in table \ref{XGBoost Hyperparameter}.
\begin{table}[ht]
    \centering
    \caption{XGBoost: Selected Hyperparameter for Model for Europe}
    \begin{tabular}{|c@{}|c@{}|c@{}|c@{}|c@{}|c@{}|}
    \hline 
          &pH\_CaCl2&pH\_H2O&Phosphorus&Nitrogen&Potassium\\
    \hline
         max\_depth & 9 & 8 & 4 & 6 & 3 \\
         learning\_rate & 0.2413 & 0.2380 & 0.2398 & 0.1945 & 0.2573 \\
         subsample & 0.8793 & 0.8366 & 0.8571 & 0.6997 & 0.6758 \\
         colsample\_bytree & 0.8405 & 0.8939 & 0.6542 & 0.7399 & 0.6788 \\
         gamma & 0.7073 & 0.4134 & 0.8846 & 0.5221 & 0.5713 \\
         reg\_alpha & 0.3075 & 0.9372 & 0.6006 & 0.4994 & 0.5544 \\
         reg\_lambda & 0.7443 & 0.0988 & 0.5232 & 0.9292 & 0.5435 \\
    \hline
    \end{tabular}
    \label{XGBoost Hyperparameter}
\end{table}
These optimized parameters highlight that model complexity and regularization has to be tailored to the specific characteristics of the target nutrition. The models for pH have deeper trees (8 and 9), indicating that predicting pH values requires greater model complexity- possibly because pH is influenced by multiple non-linear interactions. Phosphorus, nitrogen, and potassium, on the other hand, have moderate values, suggesting a balanced regularization approach.
To assess the model performances of the trained XGBoost models, the test dataset is applied to the models to calculate the RMSE. The results are presented in table \ref{ModelAResults}.\\

\subsubsection[A-FCNN]{Fully Connected Neuronal Network}
The FCNN serves as a foundational architecture in this study, leveraging multiple hidden layers to capture complex relationships in the data. To ensure an optimal model configuration, a hyperparameter search was conducted using Optuna, allowing for dynamic adjustments based on performance criteria. Various architectural and training parameters were explored, including the number of hidden layers, neurons per layer, dropout rates, learning rates, optimizers, and batch sizes. These parameters are fine-tuned to balance model complexity and generalization, minimizing the risk of overfitting while ensuring efficient learning. The key hyperparameter settings are detailed as follows:
\begin{itemize}
    \item Number of Hidden Layers: The network depth is optimized over a range of 1 to 5 layers.
    \item Number of Neurons per Layer: For each hidden layer, the number of neurons is determined individually within the range of 8 to 128 (step size 4).
    \item Dropout Rate: To avoid overfitting, the dropout functionality is used with a rate between 0.1 and 0.5.
    \item Learning Rate: The learning rate is varied between 0.0001 and 0.01 to control model adaptation.
    \item Optimizer: Only SGD and Adam are considered in this project.
    \item Batch Size: Three options were evaluated—16, 32, or 64.
\end{itemize}
The selected hyperparameters vary across models to optimize performance, as shown in Table \ref{NN Hyperparameter}. The number of hidden layers ranges from 8 to 18, with Adam as the chosen optimizer for all models. Learning rates are adjusted per target variable, with phosphorus requiring the lowest (0.00057) and pH in $CaCl_{2}$ the highest (0.00568). Batch sizes differ, with 16 or 32 for pH models and 64 for nutrients. These variations highlight the need for tailored configurations to ensure optimal training and generalization.\\
\begin{table}[ht]
    \centering
    \caption{FCNN: Selected Hyperparameter for Model for Europe}
    \begin{tabular}{|c@{}|c@{}|c@{}|c@{}|c@{}|c@{}|}
    \hline 
          &pH\_CaCl2&pH\_H2O&Phosphorus&Nitrogen&Potassium\\
    \hline
         \# hidden\_layer & 13 & 8 & 9 & 18 & 8 \\
         learning\_rate & 0.00568 & 0.00031 & 0.00057 & 0.00164 & 0.00136 \\
         optimizer & Adam & Adam & Adam & Adam & Adam \\
         batch\_size & 16 & 32 & 64 & 64 & 64 \\
    \hline
    \end{tabular}
    \label{NN Hyperparameter}
\end{table}
 
\subsubsection[A-RF]{Random Forest}
As for the other models, optuna is used to optimize the hyperparameters for determining the best performing model.

The following hyperparameters are optimized during the hyperparameter tuning process:
\begin{itemize}
    \item Number of Estimators: A range of 50 to 500 is given to select the number of estimators.
    \item Max Depth: The value for the maximum depth of the decision trees is optimized in the range from 3 to 30.
    \item Minimum Sample Split: The minimum sample split starts at 2 and ends up to 20.
    \item Minimum Sample Leafs: The minimum sample leafs starts at 1 and ends up to 20.
    \item Maximum Features: The maximum feature range is between 0.1 to 1.0.
\end{itemize}
An overview of the selected random forest hyperparameters for each target value is shown in table \ref{RF Hyperparameter}.
\begin{table}[ht]
    \centering
    \caption{Random Forest: Selected Hyperparameter for Model for Europe}
    \begin{tabular}{|@{}c@{}|c@{}|c@{}|c@{}|c@{}|c@{}|}
    \hline 
          &pH\_CaCl2&pH\_H2O&Phosphorus&Nitrogen&Potassium\\
    \hline
         estimators & 352 & 447 & 402 & 451 & 331 \\
         max\_depth & 14 & 17 & 10 & 9 & 19 \\
         min\_samples\_split & 14 & 17 & 7 & 14 & 15 \\
         min\_samples\_leaf & 6 & 7 & 11 & 10 & 17 \\
         max\_features & 0.7323 & 0.4630 & 0.7243 & 0.9900 & 0.1101 \\
    \hline
    \end{tabular}
    \label{RF Hyperparameter}
\end{table}

\subsection[A-Results]{Results}
This chapter presents the achieved model performance and discusses feature importance. The analysis is done for the five key soil parameters: pH (measured in $CaCl_{2}$ and $H_{2}O$), phosphorus, nitrogen, and potassium.\\

\subsubsection[A-Model-Performance]{Model Performance}
In the following section, the results obtained by the three models are presented. As a performance metric, the Root Mean Squared Error (RMSE) of the final models is calculated and displayed in Table \ref{ModelAResults}. The RMSE is determined using the test dataset from the single-split method described in Section \ref{SectionSingleSplit}. Each of the three model variants is trained for each of the five nutrients.
\begin{table}[ht]
    \centering
    \caption{Model for Europe: Performance Results Test Dataset}
    \begin{tabular}{|c@{}|c@{}|c@{}|c@{}|c@{}|}
    \hline
    \textbf{Model}&\textbf{Nutrient}&\textbf{Unit}&\textbf{(Mean ± StdDev)}&\textbf{RMSE}\\
    \textbf{Variant}&               &             &                       & \small (Test)   \\
    \hline
    XGBoost & pH in CaCl2 & - & 5.71 ± 1.40&1.09\\ 
            & pH in H2O & - & 6.26 ± 1.32&1.03 \\ 
            & Phosphorus, extractable (P) & mg/kg & 26.95 ± 27.02&26.53 \\ 
            & Nitrogen, extractable (N) & g/kg & 3.15 ± 3.70&3.63 \\
            & Potassium, extractable (K) & mg/kg &	204.83 ± 208.25&216.48 \\ \hline
    FCNN & pH in CaCl2  & - & 5.71 ± 1.40&1.12 \\ 
            & pH in H2O & - & 6.26 ± 1.32&1.08 \\ 
            & Phosphorus, extractable (P) & mg/kg & 26.95 ± 27.02&25.50 \\ 
            & Nitrogen, extractable (N) & g/kg & 3.15 ± 3.70&3.44 \\ 
            & Potassium, extractable (K) & mg/kg & 204.83 ± 208.25&178.20 \\ \hline
    Random  & pH in CaCl2  & - &5.71 ± 1.40&1.09 \\ 
    Forest  & pH in H2O & - & 6.26 ± 1.32&1.02 \\ 
            & Phosphorus, extractable (P) & mg/kg & 26.95 ± 27.02&26.50 \\ 
            & Nitrogen, extractable (N) & g/kg & 3.15 ± 3.70&3.63 \\ 
            & Potassium, extractable (K) & mg/kg & 204.83 ± 208.25&216.06 \\ \hline
    \end{tabular}
    \label{ModelAResults}
\end{table}

\paragraph[A-Model-Performance-XGBoost]{Results XGBoost}\label{Chapter03_ResXgboost}
When comparing the RMSE for pH in $CaCl_2$ and pH in $H_{2}O$, similar values are observed, indicating comparable accuracy in predicting pH values in both media. Because the pH range extends from 0 to 14, these errors are regarded as moderate.
The RMSE of 26.53mg/kg for phosphorus and 3.15g/kg for nitrogen are both close to the standard deviations within the dataset, suggesting that the prediction accuracy for these nutrients aligns well with the observed variability but could still be improved. The RMSE of 216.48mg/kg for potassium is higher yet remains close to the dataset’s standard deviation (208.25mg/kg), indicating that the XGBoost model captures variations in potassium levels reasonably well.
Additionally, a comparison between the single-split and spatial CV approaches is conducted for the XGBoost model. The model is trained using both the single-split approach (with separate training and test datasets) and spatial CV (with training, validation, and test datasets). The results of this comparison are displayed in Table \ref{ResultsSCV}.
\begin{table}[ht]
    \centering
    \caption{Comparison XGBoost Results: Single Split vs. Spatial CV}
    \begin{tabular}{|c|c|c|c@{}|c@{}|}
    \hline
    \textbf{Nutrient} & \textbf{Unit} & \textbf{RMSE$_{Test}$} & \multicolumn{2}{c|}{\textbf{Spatial CV}} \\ 
    &   & \textbf{(Single Split)} & \textbf{RMSE$_{Average}$} & \textbf{RMSE$_{Test}$} \\ \hline
    pH in CaCl2 & - &  1.09 & 1.09 & 1.15 \\ \hline
    pH in H2O & -  & 1.03 & 1.03 & 1.10 \\ \hline
    Phosphorus & mg/kg &  26.53 & 26.98 & 26.32 \\ \hline
    Nitrogen & g/kg &  3.63 & 3.72 & 2.46 \\ \hline
    Potassium & mg/kg & 216.48 & 207.68 & 177.52 \\ \hline
    \end{tabular}
    \label{ResultsSCV}
\end{table}
In column three, the results of the single-split method, described earlier in this section, are presented for comparison. Columns four and five show the metrics RMSE\textsubscript{Average} and RMSE\textsubscript{Test} for the spatial CV. The results for the two pH values exhibit only minor differences between the single split and spatial CV, suggesting that no significant geographical dependencies exist for pH predictions. A similar observation applies to phosphorus, where model performance does not notably differ across the two methods.
A more pronounced difference is observed when nitrogen and potassium results are compared between the single split and spatial CV. The RMSE\textsubscript{Test} for the spatial CV is lower than for the single-split method, which is unusual for spatial CV. Analysis indicates that the distribution of data points in the spatial CV test dataset is more favorable for the model, with many test data points resembling those in the training set and containing fewer extreme or rare values. Consequently, the lower RMSE\textsubscript{Test} may be misleading, as the test dataset is not fully representative of the overall data distribution. This issue arises because the test set in spatial CV is defined by geographic grids rather than random selection, potentially reducing the diversity of cases included in the test subset.
Since the XGBoost-based analysis does not reveal substantial differences between the two validation methods (except for nitrogen and potassium), the remaining models are trained using the less complex single-split method. Nevertheless, spatial CV remains a promising avenue for future work. The aforementioned limitations of grid-based spatial CV should be taken into account, and more sophisticated approaches to data partitioning should be explored.

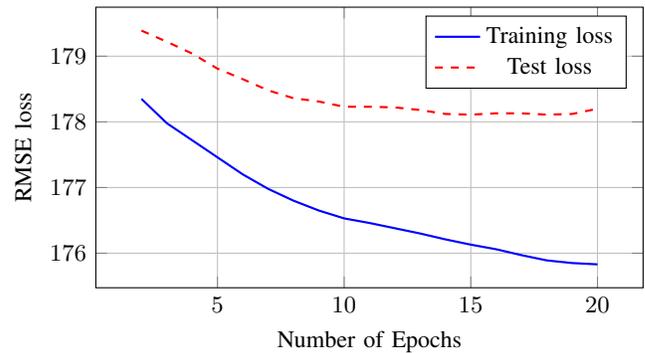
\begin{figure}[ht]
    \centering
    \begin{tikzpicture}
    \begin{axis}[
        xlabel={Number of Epochs},
        ylabel={RMSE loss},
        grid=major,
        width=1.0\columnwidth,
        height=0.6\columnwidth,
        legend pos=north east,
        legend style={font=\small},
        label style={font=\small},
        tick label style={font=\small}
    ]
    \addplot[color=blue, thick] coordinates {
        (2,178.35)
        (3,177.98)
        (4,177.72)
        (5,177.46)
        (6,177.20)
        (7,176.98)
        (8,176.80)
        (9,176.65)
        (10,176.53)
        (11,176.46)
        (12,176.38)
        (13,176.30)
        (14,176.21)
        (15,176.13)
        (16,176.06)
        (17,175.97)
        (18,175.89)
        (19,175.85)
        (20,175.83)
    };
    \addlegendentry{Training loss}
    \addplot[color=red, dashed, thick] coordinates {
        (2,179.39)
        (3,179.22)
        (4,179.04)
        (5,178.81)
        (6,178.65)
        (7,178.48)
        (8,178.36)
        (9,178.31)
        (10,178.23)
        (11,178.23)
        (12,178.22)
        (13,178.18)
        (14,178.12)
        (15,178.11)
        (16,178.13)
        (17,178.13)
        (18,178.11)
        (19,178.12)
        (20,178.20)
    };
    \addlegendentry{Test loss}
    \end{axis}
    \end{tikzpicture}
    \caption{Compare Training and Test loss of potassium}
    \label{fig:training-test-loss}
\end{figure}

\paragraph[A-Model-PerformanceFCNN]{Results FCNN}
The FCNN initially underperforms compared to XGBoost and Random Forest in predicting phosphorus, with an RMSE of 27.12, while XGBoost achieves 26.53. Investigation into the model's performance indicates potential overfitting or underfitting issues. Despite attempts to mitigate overfitting through regularization, improvements remain marginal. However, an intermediate FCNN model with two hidden layers and 152 neurons unexpectedly surpasses XGBoost, achieving an RMSE of 26.22. Further optimization using Optuna, with the number of hidden layers restricted between three and nine, yields an optimized FCNN featuring nine hidden layers and 708 neurons, which achieves a notably lower RMSE of 25.50.
The final FCNN model also outperforms XGBoost and Random Forest in predicting nitrogen and potassium, with RMSE values of 3.44 and 178.20, respectively, compared to XGBoost's 3.63 and 216.48. To verify generalization, training and test loss trends are examined. The initial epoch (error = 198.13) is excluded for clarity, revealing a consistent decline in both losses over the first ten epochs. From epoch 15 onward, the test loss remains stable while training loss continues to drop, suggesting the onset of overfitting. Figure~\ref{fig:training-test-loss} illustrates the loss progression, confirming the model’s robustness.

\paragraph[A-Model-Performance-RF]{Results Random Forest}
The performance results of the Random Forest model are presented in Table \ref{ModelAResults}. For pH measurements, the model achieves a RMSE of 1.09 for pH in $CaCl_2$ and 1.02 for pH in $H_{2}O$, demonstrating relatively low error values. Regarding phosphorus, the RMSE is 26.50 mg/kg, closely aligning with the mean and standard deviation of the dataset, suggesting a moderate predictive performance. Similarly, for nitrogen, the model produces an RMSE of 3.63 g/kg, which is comparable to its standard deviation, indicating a consistent error margin. The highest RMSE is observed for potassium, with a value of 216.06 mg/kg, reflecting a greater variability in the dataset. Overall, the Random Forest model provides competitive results, particularly for pH prediction, while showing larger errors in nutrient predictions, likely due to the high variability within the dataset.

\paragraph[A-Model-Performance-RF]{Comparison of Results}
The performance evaluation of the three model types, XGBoost, FCNN, and Random Forest, highlights their predictive capabilities across different soil nutrients (Table \ref{ModelAResults}). XGBoost demonstrates strong overall performance, achieving an RMSE of 26.53mg/kg for phosphorus, 3.63 g/kg for nitrogen, and 216.48mg/kg for potassium. Its RMSE for pH prediction remains low, with values of 1.09 for pH in $CaCl_{2}$ and 1.03 for pH in $H_{2}O$, indicating a reliable predictive ability for soil acidity. While XGBoost performs well for pH and phosphorus, its errors for nitrogen and potassium are comparable to those of Random Forest, suggesting challenges in capturing variability in nutrient concentrations.
The Random Forest model delivers similar performance, particularly for pH prediction, with RMSE values of 1.09 and 1.02 for pH in $CaCl_{2}$ and $H_{2}O$, respectively. For phosphorus and nitrogen, it achieves RMSE values of 26.50mg/kg and 3.63g/kg, closely aligning with those of XGBoost. However, its largest error is observed for potassium, with an RMSE of 216.06mg/kg, reflecting high variability within the dataset. Despite competitive performance, especially for pH prediction, its accuracy in nutrient estimation remains limited.\\
In contrast, the optimized FCNN model surpasses both XGBoost and Random Forest for phosphorus, nitrogen, and potassium prediction. After hyperparameter tuning using Optuna, the FCNN achieves an RMSE of 25.50mg/kg for phosphorus, improving upon XGBoost and Random Forest. Additionally, it reduces the RMSE for nitrogen to 3.44g/kg and for potassium to 178.20mg/kg, outperforming the other models. pH predictions, however, remain slightly less accurate, with RMSE values of 1.12 for pH in $CaCl_{2}$ and 1.08 for pH in $H_{2}O$. Training and test loss evaluations confirm the model’s robustness, as illustrated in Figure \ref{fig:training-test-loss}. Overall, the FCNN demonstrates superior performance for nutrient prediction, particularly for phosphorus and potassium, while maintaining competitive accuracy for pH measurements.\\

\subsubsection[A-Feature-Importance]{Feature Importance}
\begin{figure}[htb]
    \centering
    \includegraphics[width=1\linewidth]{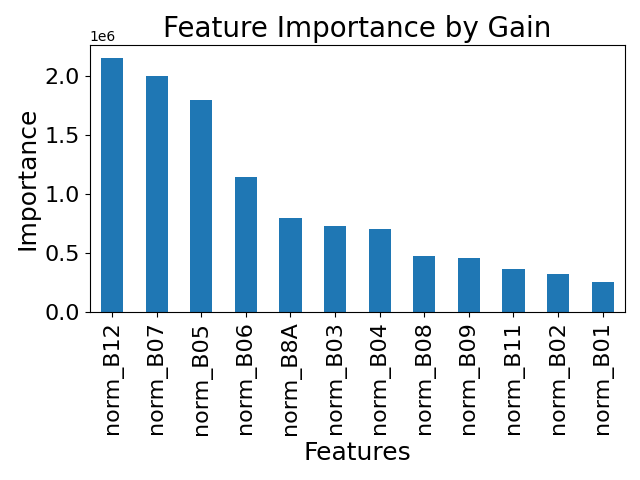}
    \caption{Feature Importance for K - XGBoost Model for Europe}
    \label{fig:03_FeatImp_A_K}
\end{figure}

\begin{figure}[htb]
    \centering
    \includegraphics[width=1\linewidth]{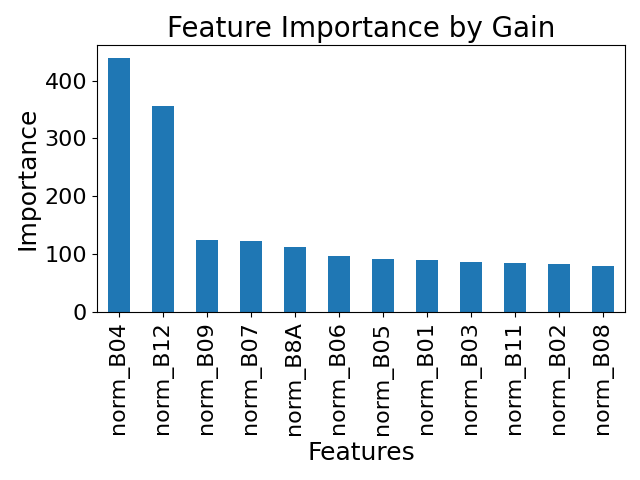}
    \caption{Feature Importance for N - XGBoost Model for Europe}
    \label{fig:03_FeatImp_A_N}
\end{figure}

\begin{figure}[htb]
    \centering
    \includegraphics[width=1\linewidth]{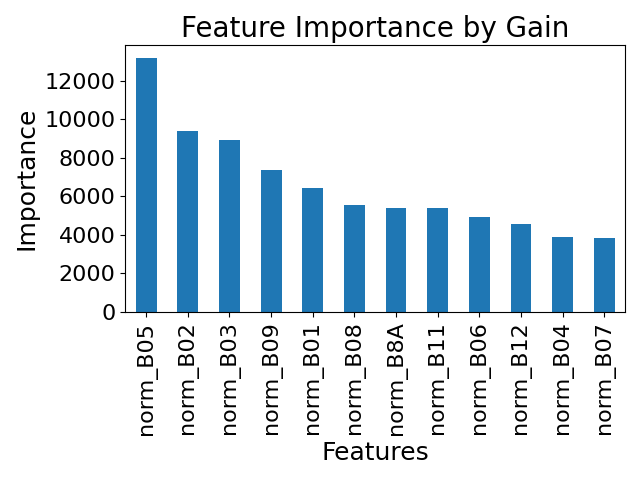}
    \caption{Feature Importance for P - XGBoost Model for Europe}
    \label{fig:03_FeatImp_A_P}
\end{figure}

\begin{figure}[htb]
    \centering
    \includegraphics[width=1\linewidth]{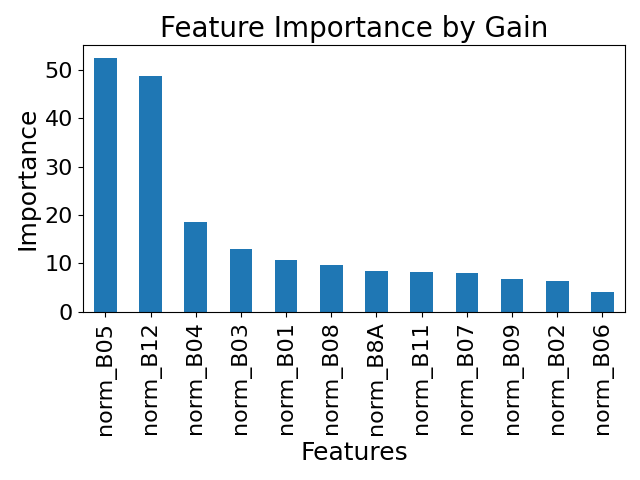}
    \caption{Feature Importance for pH from H2O - XGBoost Model for Europe}
    \label{fig:03_FeatImp_A_H2O}
\end{figure}

\begin{figure}[htb]
    \centering
    \includegraphics[width=1\linewidth]{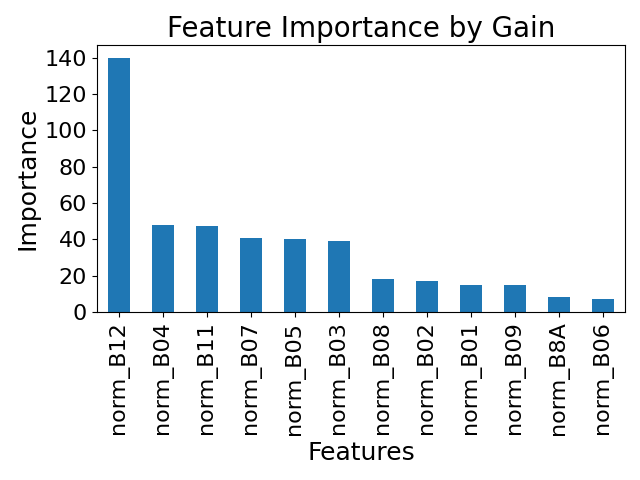}
    \caption{Feature Importance for pH from CaCl2 - XGBoost Model for Europe}
    \label{fig:03_FeatImp_A_CaCl2}
\end{figure}

In the following, feature importance is illustrated for the XGBoost model. The evaluation is based on the average gain across all splits in which a given feature is utilized.
For potassium (see Figure \ref{fig:03_FeatImp_A_K}), there is no single dominant feature. Instead, three Sentinel-2 bands (B12, B07, and B05) emerge as the most influential for predicting this target variable.
In figure \ref{fig:03_FeatImp_A_N} the most important features for nitrogen are the bands B04 and B12. All other bands have a minor contribution.
For phosphorus, the XGBoost feature importance illustrated in Figure \ref{fig:03_FeatImp_A_P} indicates that band B05 has the greatest influence, while bands B02 and B03 also contribute significantly to the model.
Finally, the analysis for pH reveals some notable observations. Because two target values are available—one measured in $H_{2}O$ and another in $CaCl_2$—it is possible to compare feature importance for each measurement method (Figures \ref{fig:03_FeatImp_A_H2O} and \ref{fig:03_FeatImp_A_CaCl2}). In both cases, B12 emerges as a highly influential feature, serving as the most important feature for $CaCl_2$ and ranking second for $H_{2}O$. For pH measured in $H_{2}O$, band B05 takes on a slightly higher importance. Although this may seem to conflict with the high linear correlation between pH in $H_{2}O$ and $CaCl_2$, as detailed in Paragraph \ref{Europe-Corr-Input}, these differences in feature importance are not surprising given the non-linear and multi-split nature of tree-based models.

To summarize the feature importance results, band 12 is identified as the most influential feature for the model's predictions. In four out of the five cases, band 12 is ranked among the top two features—it is the most important feature in two cases and the second most important in another two, with the only exception being the phosphorus prediction where it plays a minor role. Overall, band 05 emerges as the next most influential feature, achieving the top rank twice.

\section[A*]{Extended Model for Europe}

Based on the Model for Europe in the previous section \ref{modeleurope}, we decided to extend the scope to train the model by using additional features. We therefore expect increased precision and reduced prediction error with this extended model compared to the previous, simpler input model.

\subsection[A*-Data-Set]{Data Set}
In addition to the single pixel from Sentinel-2 satellite image bands, we consider the following data as additional features for model training.

\begin{itemize}
\item 8 neighbor pixels: 3 x 3 = 9 pixels instead of the single pixel
\item Weather data: 9 features
\item Crop yield scores: 27 features
\item Clay model embeddings from Masked Auto Encoder (MAE): 1024 features\\
\end{itemize}

\subsubsection{Neighbor Pixels of the Sentinel Data}
\begin{figure}[ht]
    \centering
    \includegraphics[width=1.0\linewidth]{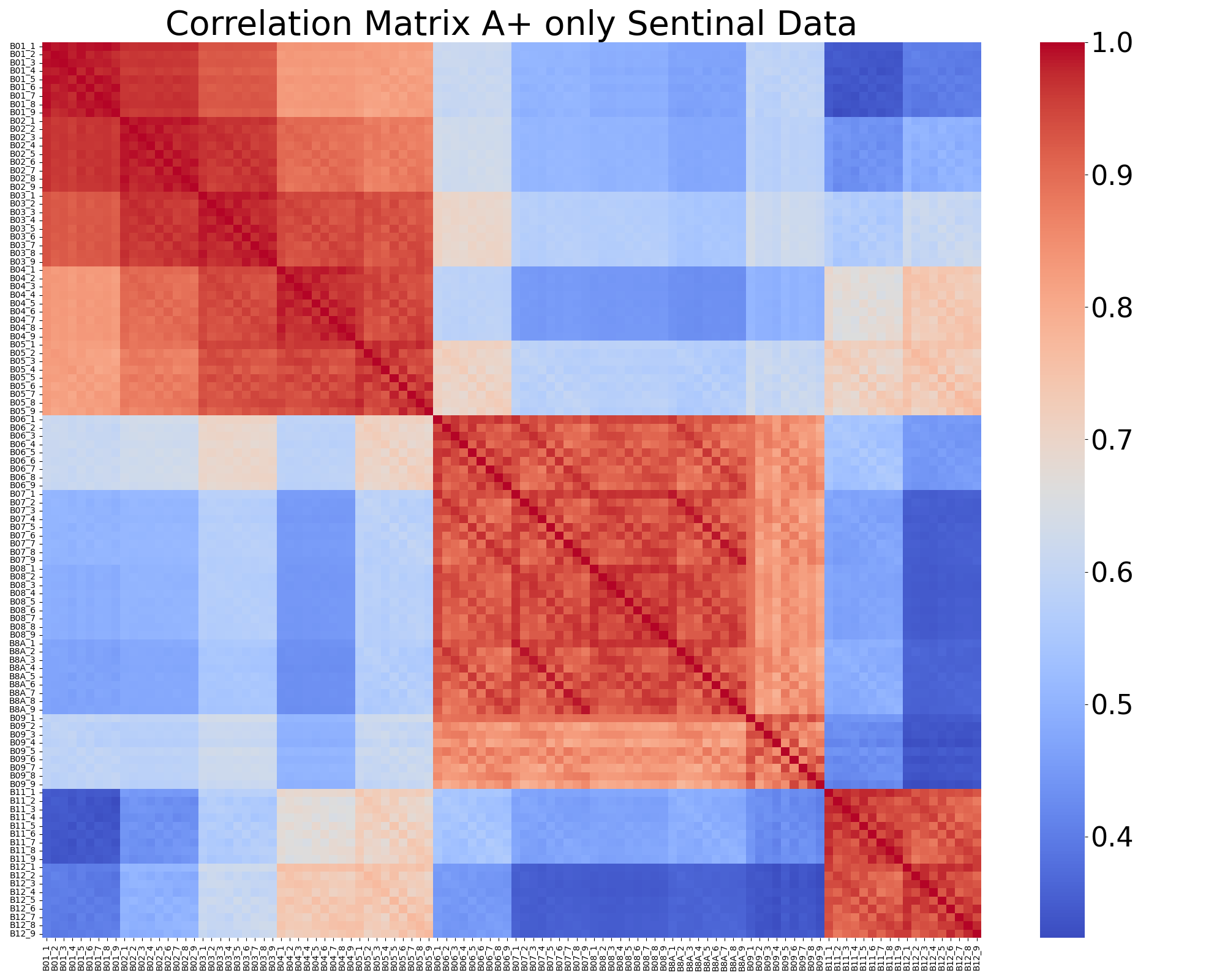}
    \caption{Correlation coefficient matrix for the neighbor pixel of the Sentinel bands for the extended Europe model}
    \label{fig:04_Sentinel-Neighbor_corr}
\end{figure}

Satellite data are, by nature, image data. For ML tasks involving images, neighboring pixels are often used—such as in Convolutional Neural Networks—to capture local spatial correlations \cite{Goodfellow-et-al-2016}. In a soil prediction task, the use of neighbor pixels must be balanced between more neighbor pixels to get more information and less neighbor pixels to reduce unwanted noise. This noise occurs due to the large size of a pixel (10 m x 10 m up to 60m x 60m) compared with the small size of a field in most areas of the world. Using too many pixels can increase noise from adjacent buildings, streets, and non-agricultural areas. Therefore only direct neighbor pixels are used.\\
An investigation of the linear correlation coefficient matrix shown in figure \ref{fig:04_Sentinel-Neighbor_corr}, shows the very similar fundamental correlation between bands like in figure \ref{fig:03-CorrCoeff-Input}, but with a higher "resolution". This is caused by the very high correlation of the additional neighbor pixels to the central pixel. Given that a change in information over a spatial step of 10 m to 60 m is unlikely, it is inferred that the additional neighbor pixels do not provide much extra information to the model.\\

\subsubsection[Weather Data]{Weather Data}
Weather data is integral to understanding soil nutrient dynamics, as climatic variables strongly influence moisture availability and nutrient leaching. In this project, weather data (e.g., temperature, precipitation, dew point) is aligned with the satellite image timestamps to ensure that each soil sample’s spectral reading corresponds to concurrent weather conditions. Access to this historical data incurs costs after 1,000 API calls per day.\\

\begin{figure}[ht]
    \centering
    \includegraphics[width=1.0\linewidth]{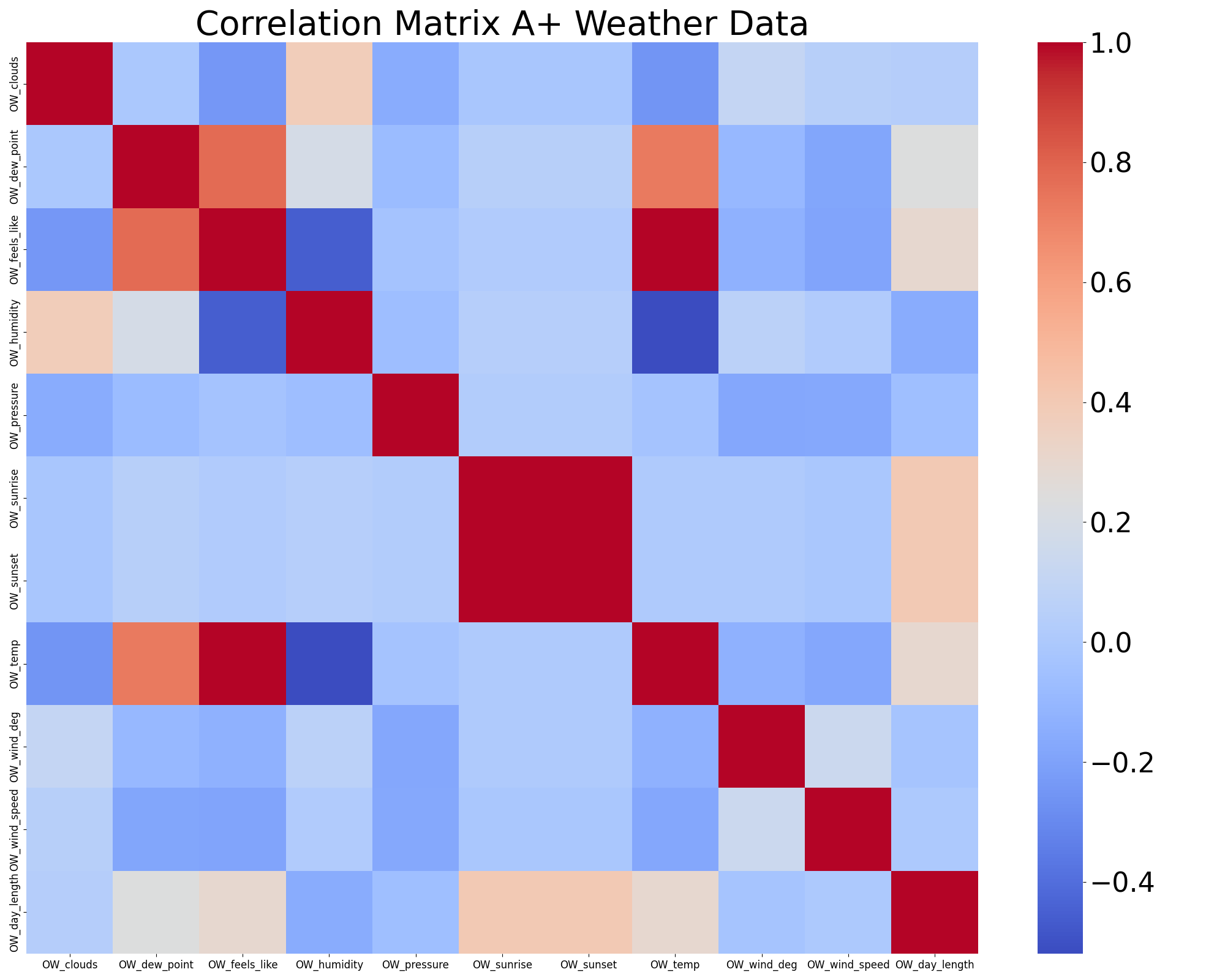}
    \caption{Correlation coefficient matrix for the weather data for the extended Europe model}
    \label{fig:04_weather_corr}
\end{figure}

Figure \ref{fig:04_weather_corr} presents the correlation coefficient matrix for the extracted weather variables. As expected, temperature (\texttt{OW\_temp}) shows a nearly perfect correlation with the temperature-feels-like metric (\texttt{OW\_feels\_like}). Additionally, dew point (\texttt{OW\_dew\_point}) is highly correlated with both \texttt{OW\_temp} and \texttt{OW\_feels\_like}, illustrating their shared underlying physical principles.

While such high correlations can reduce feature diversity, they also reaffirm that these variables capture significant climate-driven effects that potentially influence topsoil properties.\\
By incorporating these weather variables into the extended model, a modest improvement in RMSE scores was observed for certain nutrients (e.g., phosphorus and potassium). This improvement suggests that specific climatic conditions—such as rainfall patterns or heat stress—can provide meaningful context beyond raw reflectance signals alone.\\

\subsubsection[Crop Yield Scores]{Crop Yield Scores}
Crop yield data serves as a proxy for soil fertility when direct soil measurements are either unavailable or incomplete. In this study, the project team integrated yield metrics derived from the FAO GAEZ portal, which provides production estimates for various crops (e.g., wheat, maize, barley, and oilseeds) across different regions. These yield values, normalized to ensure comparability, were aligned with the spatial coordinates of soil sampling.

\paragraph{Rationale for Including Yield Data}
High or low crop yields typically reflect underlying soil characteristics, climatic conditions, and farm management practices. By incorporating crop yield information, the model gains an indirect measure of long-term soil productivity, bridging potential gaps in direct soil property data. Furthermore, it provides additional context information for satellite images since it may reveal characteristic crop reflectance patterns at the pixel level.

\begin{figure}[ht]
    \centering
    \includegraphics[width=1.0\linewidth]{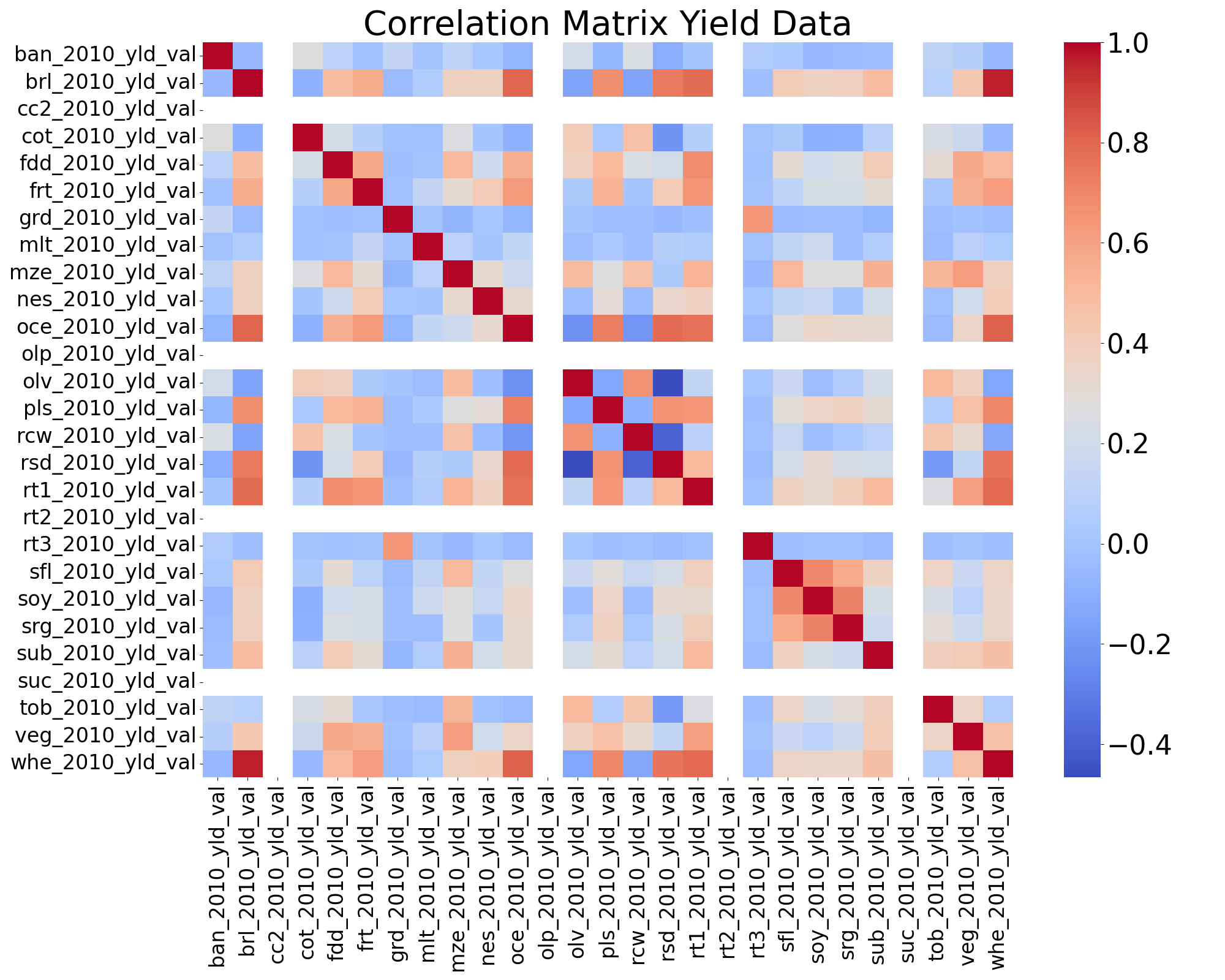}
    \caption{Correlation coefficient matrix for the World Yield data for the extended Europe model}
    \label{fig:04_Yield-Real_corr}
\end{figure}

\paragraph{Correlation Analysis}
A correlation matrix of the yield variables (Figure~\ref{fig:04_Yield-Real_corr}) revealed that certain crops share strong linear relationships (for instance, barley (\texttt{brl}) and wheat (\texttt{whe})), reflecting similar agronomic requirements. Meanwhile, crops not commonly cultivated in Europe lacked data points in this region, leading to sparse or zero values. Despite these gaps, the inclusion of yield scores contributed to capturing broader agricultural contexts, thereby enhancing model performance for certain nutrients (notably nitrogen and phosphorus). However, excessive reliance on yield data could introduce biases if confounding factors (e.g., irrigation, fertilization regimes) are not adequately represented.\\

\subsubsection[A*-Clay-Model]{Clay Model}

The Clay model is an open-source foundation model for Earth observation \cite{clay2024}. The goal of Clay is to make data cheaper, easier to use, and more accessible to communities and everyone working on climate and nature. Clay’s model uses satellite imagery, as well as information about location (longitude, latitude) and time, as input. As output we obtain embeddings, which can be considered as mathematical representations of a given area at a certain time on the Earth's surface. Vision Transformer architecture is implemented for understanding geospatial and temporal relations on Earth Observation data. A Masked Autoencoder is used as a self-supervised learning method to train the models. In our project, we decided to use a pre-trained Clay model to generate semantic embeddings as extra features for training our own ML models for the soil prediction.
 
The Clay model expects a dictionary with the following keys of certain dimensions \cite{clay2024}:
\begin{itemize}
\item pixels: batch x band x height x width - normalized values
\item time: batch x 4 - horizontally stacked week\textunderscore norm and hour\textunderscore norm
\item latlon: batch x 4 - horizontally stacked lat\textunderscore norm and lon\textunderscore norm
\item waves: list[:band] - wavelength of each band of the sensor from sentinel-2
\item gsd: scalar - ground sample distance of the sensor from sentinel-2
\end{itemize}

The raw data will be preprocessed into the required dimension of the input data as listed above. Clay expects normalized pixel data coming from the satellite images. 9 x 9 pixels of each band are used as input. We set all the values of the parameter of the key "time" to zero, since we are much more interested in the location of the soil data and we don't transform the timestamp to weeks and hours as Clay requires. The values of longitude and latitude are normed using sine and cosine functions, respectively. That is the reason, why 4 values are needed for the key "latlon" for each data set. The ground sample distance is 60 m. It is pre-defined in order to use information of all the available 12 bands of the Sentinel-2 satellite.

The checkpoint we use, in which the pre-trained weights and bias are stored, can be downloaded from: \href{https://huggingface.co/made-with-clay/Clay/resolve/main/v1.5/clay-v1.5.ckpt}{https://huggingface.co/made-with-clay/Clay/resolve/main/v1.5/clay-v1.5.ckpt}. The Clay model itself offers different model sizes as options: "tiny", "small", "base", and "large". The parameters from the checkpoint were trained with the "large" model, which uses a kernel size of 8 x 8. Decision has to be made by choosing the pixel resolution of the satellite images, so that the "large" model can be applied to the data set. Usage of other size of models leads to mismatching errors, because the dimension of weights and bias from the pre-trained model checkpoint and the current model don't match.

After preprocessing, we have a dictionary with the following keys and dimensions:
\begin{itemize}
\item pixels: batch x 12 x 9 x 9 
\item time: batch x 4 
\item latlon: batch x 4 
\item waves: list[:band] 1 x 12
\item gsd: 60 
\end{itemize}

\begin{figure}[ht]
    \centering
    \includegraphics[width=0.9\linewidth]{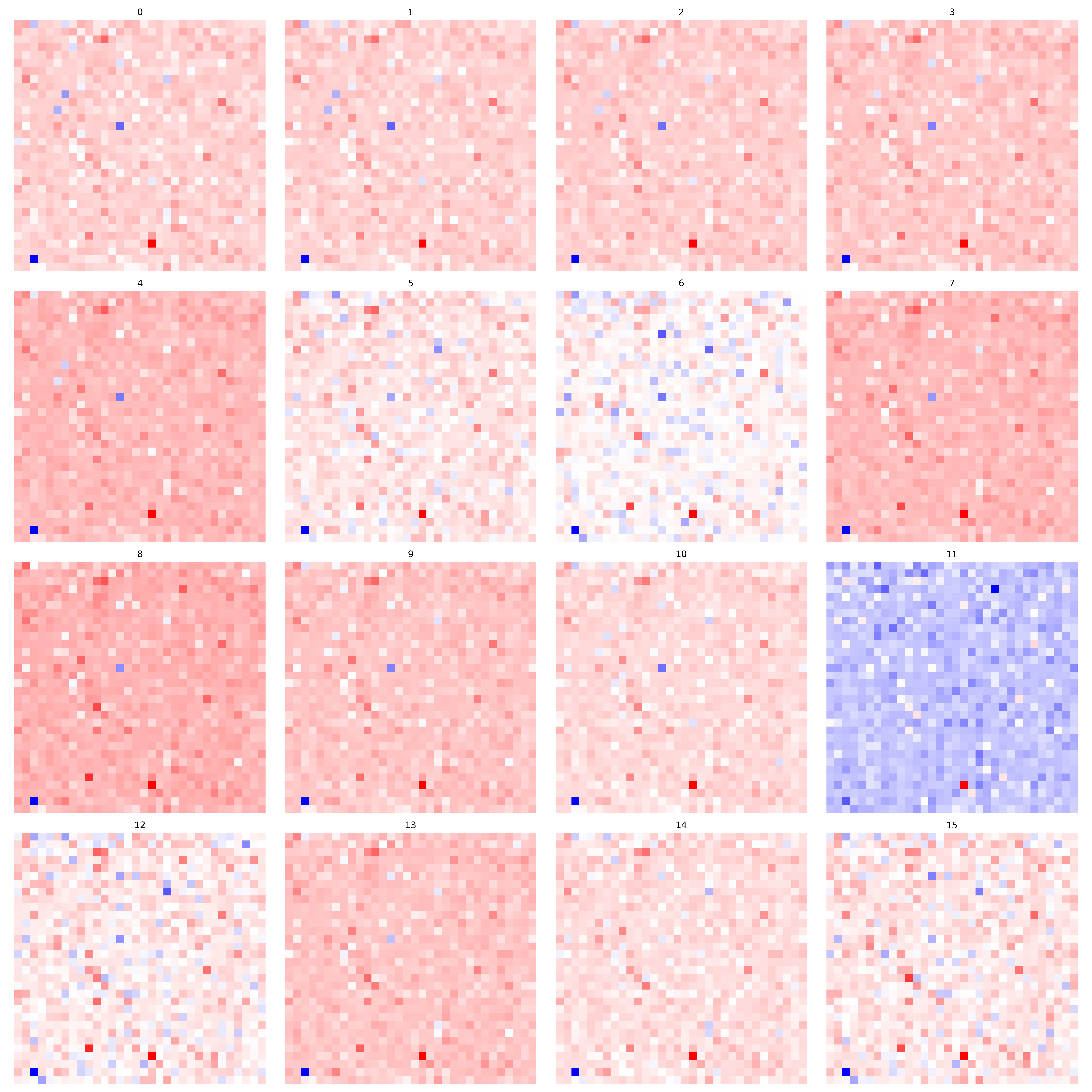}
    \caption{Visualization of Clay embeddings}
    \label{fig:clay}
\end{figure}

A data cube of all the keys and values as described above is prepared to feed the Clay model. The Clay model is imported to generate the embeddings with a dimension of 1024. It means that additional 1024 features are created to train our own models for soil prediction. For the illustration, the embeddings of the first 16 data sets are plotted as shown in figure \ref{fig:clay} using a "bwr" (blue-white-red) color map. We transformed 1024 to a 32 x 32 matrix for better visualization.

\begin{figure}[ht]
    \centering
    \includegraphics[width=1.0\linewidth]{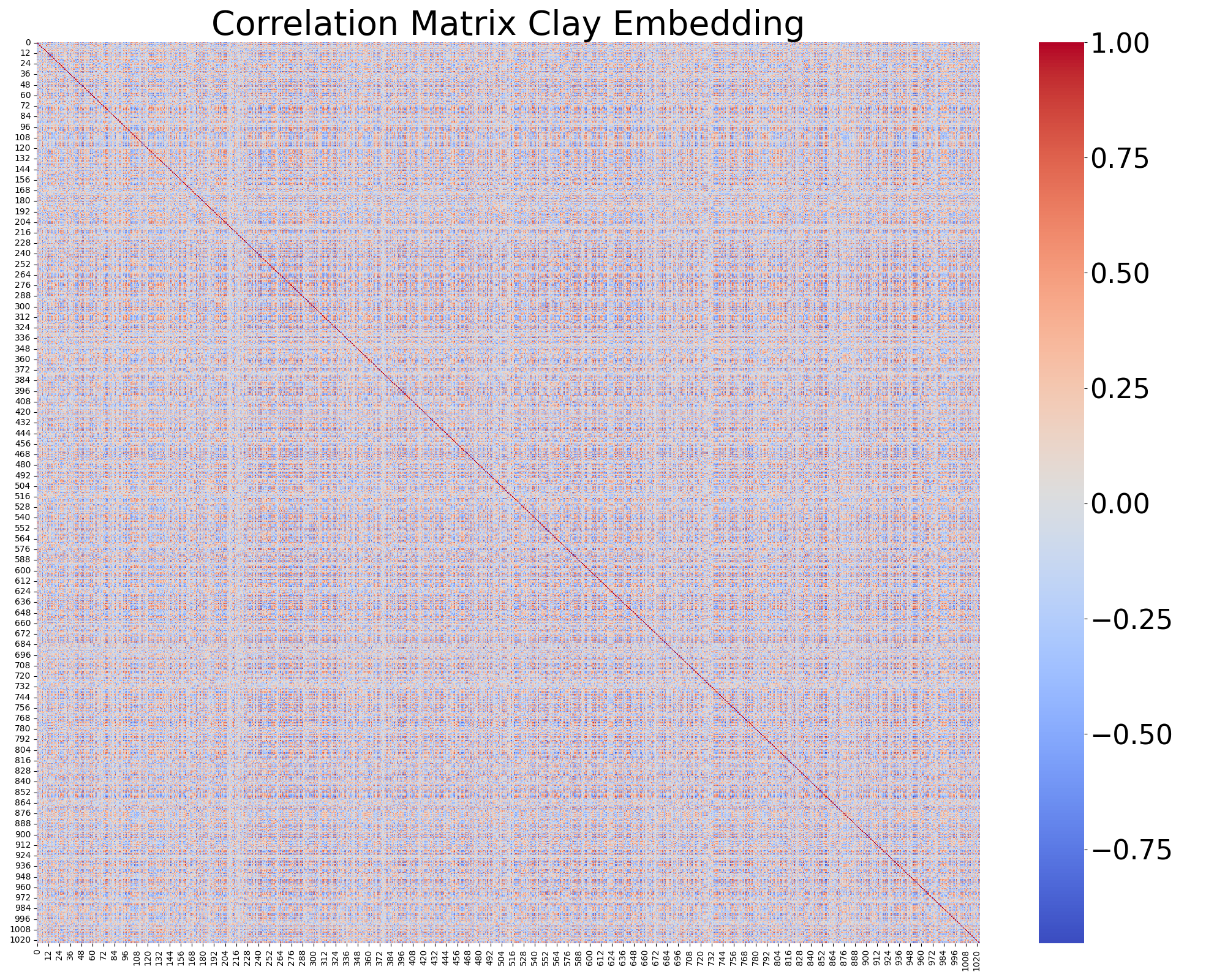}
    \caption{Correlation coefficient matrix for the Clay embeddings for the extended Europe model}
    \label{fig:04_clay_corr}
\end{figure}

To complete the analysis of the Clay embeddings the linear correlation coefficients are shown  in figure \ref{fig:04_clay_corr}. As to expect for the abstract embeddings, no tendency towards a high positive or negative correlation can be observed.\\

\subsubsection{Data Combinations for Training}
\label{sec:data-combinations}

Because multiple data sources are introduced in the extended model, the AgroLens project evaluates several feature configurations to determine which combination delivers the most accurate soil nutrient predictions:

\begin{itemize}
    \item \textbf{Neighbor Pixels (SURR):} Expands each single-pixel input to include an adjacent $3\times3$ region, capturing local spatial variation.
    \item \textbf{Weather Data (WTHR):} Incorporates climate variables (e.g., temperature, precipitation) aligned with satellite imagery timestamps, providing environmental context.
    \item \textbf{Crop Yield Scores (CRY):} Integrates FAO yield estimates for various crops, serving as a proxy for soil productivity.
    \item \textbf{Clay Embeddings (CLAY):} Adds a 1024-dimensional representation from a pre-trained Masked Autoencoder.
\end{itemize}

Two main configurations are tested within the Extended Model for Europe:
\begin{enumerate}
    \item \textit{SURR + WTHR + CRY}: A mid-level expansion of features, hypothesized to enhance predictions by capturing local spatial detail, climate factors, and crop performance.
    \item \textit{SURR + WTHR + CRY + CLAY}: A further extension, adding the Clay embeddings to the previous set. While these embeddings can encode high-level geospatial patterns, they substantially increase the dimensionality of the feature space and resulted in slower trainable and most often less performant models.
\end{enumerate}

\subsection[A*-Model-Adaption]{Model Adaption and Training}
The primary distinction between the initial model for Europe and the extended model is an expanded set of input features. Consequently, only a change in the input vector size of the ML models is required. Furthermore, the feasibility of the parameter ranges used for hyperparameter optimization must be verified. Due to the increased volume of input data, it is possible that the models may require additional complexity to effectively process the larger amount of information. As demonstrated in the next section, \emph{Results}, the models handle the expanded input data well, although the computational time for model training increases by approximately a factor of ten.
As a quality check, the hyperparameters of the best model from the optimization run are analyzed. This analysis is performed for all models and target values. In Figure \ref{XGBoost Hyperparameter Extended}, the parameters of the best XGBoost models without Clay embedding are presented. It is observed that no hyperparameter reaches the limits of the parameter ranges provided in Section \ref{A-Used-Model-Details}, confirming that these ranges remain feasible for the extended model. Moreover, the hyperparameters vary significantly across the different target values, which further validates the decision to train separate models.

\begin{table}[ht]
    \centering
    \caption{XGBoost: Selected Hyperparameter for Extended Model without Clay Embeddings}
    \begin{tabular}{|c@{}|c@{}|c@{}|c@{}|c@{}|c@{}|}
    \hline 
          &pH\_CaCl2&pH\_H2O&Phosphorus&Nitrogen&Potassium\\
    \hline
         max\_depth        & 11     & 8      & 5      & 6      & 8 \\
         learning\_rate    & 0.2652 & 0.2505 & 0.2079 & 0.2658 & 0.2317 \\
         subsample         & 0.9767 & 0.9288 & 0.9629 & 0.8898 & 0.6218 \\
         colsample\_bytree & 0.9477 & 0.9190 & 0.9319 & 0.7511 & 0.9473 \\
         gamma             & 0.6916 & 0.9914 & 0.7509 & 0.6356 & 0.1918 \\
         reg\_alpha        & 0.5380 & 0.5489 & 0.3902 & 0.6018 & 0.3999 \\
         reg\_lambda       & 0.4387 & 0.7022 & 0.5748 & 0.1716 & 0.8859 \\
    \hline
    \end{tabular}
    \label{XGBoost Hyperparameter Extended}
\end{table}

For the FCNN, the parameters of the best model without Clay data are presented in Figure \ref{NN Hyperparameter Extended}. Significant differences between these models, as well as differences compared to the initial European model, are observed. In this comparison, the extended model without Clay requires fewer layers and a smaller learning rate than the initial European model.
\begin{table}[ht]
    \centering
    \caption{FCNN: Selected Hyperparameter for Extended Model without Clay Embeddings}
    \begin{tabular}{|c@{}|c@{}|c@{}|c@{}|c@{}|c@{}|}
    \hline 
          &pH\_CaCl2&pH\_H2O&Phosphorus&Nitrogen&Potassium\\
    \hline
         \# hidden\_layer & 8       & 9       & 4       & 4       & 7 \\
         learning\_rate   & 0.00030 & 0.00018 & 0.00019 & 0.00019 & 0.00048 \\
         optimizer        & Adam    & Adam    & Adam    & Adam    & Adam \\
         batch\_size      & 64      & 64      & 64      & 16      & 32 \\
    \hline
    \end{tabular}
    \label{NN Hyperparameter Extended}
\end{table}

Finally, for the extended model without Clay embedding using Random Forest (RF), the best parameters are shown in Figure \ref{RF Hyperparameter Extended}. Here, the same conclusions can be drawn.

\begin{table}[ht]
    \centering
    \caption{Random Forest: Selected Hyperparameter for Extended Model without Clay Embedding}
    \begin{tabular}{|@{}c@{}|c@{}|c@{}|c@{}|c@{}|c@{}|}
    \hline 
          &pH\_CaCl2&pH\_H2O&Phosphorus&Nitrogen&Potassium\\
    \hline
         estimators          & 324    & 185    & 274    & 475    & 429 \\
         max\_depth          & 16     & 19     & 23     & 27     & 23 \\
         min\_samples\_split & 19     & 15     & 17     & 6      & 15 \\
         min\_samples\_leaf  & 13     & 4      & 7      & 17     & 6 \\
         max\_features       & 0.7472 & 0.6249 & 0.4504 & 0.4401 & 0.6167 \\
    \hline
    \end{tabular}
    \label{RF Hyperparameter Extended}
\end{table}

To complete the hyperparameter analysis, the best model parameters for the input data with Clay embedding are presented. For the XGBoost model, the best parameters are shown in Figure \ref{XGBoost Hyperparameter Extended Clay}. Once again, no hyperparameter reaches the limits of the defined parameter ranges, confirming the validity of the optimization. Additionally, the parameters differ among the models for the various target values, as observed previously. In comparison to the model without Clay embedding, slight differences in the parameters are also noted.

\begin{table}[ht]
    \centering
    \caption{XGBoost: Selected Hyperparameter for Extended Model with Clay Embedding}
    \begin{tabular}{|@{}c@{}|c@{}|c@{}|c@{}|c@{}|c@{}|}
    \hline 
          &pH\_CaCl2&pH\_H2O&Phosphorus&Nitrogen&Potassium\\
    \hline
         max\_depth        & 7      & 8      & 4      & 5      & 3 \\
         learning\_rate    & 0.2352 & 0.2503 & 0.2421 & 0.1945 & 0.2705 \\
         subsample         & 0.9334 & 0.8159 & 0.9515 & 0.6997 & 0.6012 \\
         colsample\_bytree & 0.8708 & 0.8837 & 0.9161 & 0.7399 & 0.6783 \\
         gamma             & 0.0418 & 0.1945 & 0.1861 & 0.5221 & 0.1975 \\
         reg\_alpha        & 0.3183 & 0.3099 & 0.4329 & 0.4994 & 0.3617 \\
         reg\_lambda       & 0.1657 & 0.9632 & 0.0112 & 0.8032 & 0.2392 \\
    \hline
    \end{tabular}
    \label{XGBoost Hyperparameter Extended Clay}
\end{table}

Analysing the parameters for the FCNN model in figure \ref{NN Hyperparameter Extended Clay} leads to similar results.
\begin{table}[ht]
    \centering
    \caption{FCNN: Selected Hyperparameter for Extended Model with Clay Embedding}
    \begin{tabular}{|c@{}|c@{}|c@{}|c@{}|c@{}|c@{}|}
    \hline 
          &pH\_CaCl2&pH\_H2O&Phosphorus&Nitrogen&Potassium\\
    \hline
         \# hidden\_layer & 9       & 6       & 7       & 3       & 5 \\
         learning\_rate   & 0.00031 & 0.00159 & 0.00065 & 0.00042 & 0.00111 \\
         optimizer        & Adam    & Adam    & Adam    & Adam    & Adam \\
         batch\_size      & 64      & 16      & 16      & 16      & 32 \\
    \hline
    \end{tabular}
    \label{NN Hyperparameter Extended Clay}
\end{table}

The aforementioned effects are also observed for the Random Forest model parameters in Figure \ref{RF Hyperparameter Extended Clay}. In this case, the differences between models with and without Clay embedding are even more pronounced than before.
\begin{table}[ht]
    \centering
    \caption{Random Forest: Selected Hyperparameter for Extended Model with Clay Embedding}
    \begin{tabular}{|@{}c@{}|c@{}|c@{}|c@{}|c@{}|c@{}|}
    \hline 
          &pH\_CaCl2&pH\_H2O&Phosphorus&Nitrogen&Potassium\\
    \hline
         estimators          & 275    & 319    & 464    & 167    & 458 \\
         max\_depth          & 15     & 26     & 29     & 27     & 18 \\
         min\_samples\_split & 18     & 8      & 3      & 19     & 13 \\
         min\_samples\_leaf  & 3      & 2      & 17     & 15     & 20 \\
         max\_features       & 0.6019 & 0.9486 & 0.4297 & 0.1507 & 0.9337 \\
    \hline
    \end{tabular}
    \label{RF Hyperparameter Extended Clay}
\end{table}

To summarize the analysis, it is concluded that varying models for the different target values are necessary and should be considered in further developments. The differences observed for the same ML model when using different input data indicate that retraining is required whenever the input features are extended or modified. Nevertheless, the changes in the hyperparameters are not drastic, suggesting that the models are likely to yield feasible results for other features or datasets without the need for retraining.

\subsection[A*-Results]{Results}

In the following section the achieved results of the three extended models for Europe - XGBoost, FCNN and Random Forest - are presented and discussed. Table \ref{tab_extended_performance_xgboost}, \ref{tab_extended_performance_fcnn}, and \ref{tab_extended_performance_rf} each show the overall performance of their respective model for Europe (BASE) compared to the extended models for Europe. Additionally, we give an overview of the mean together with the standard deviation to get a better evaluation of RMSE values.\\

\subsubsection{XGBoost}
Table \ref{tab_extended_performance_xgboost} shows an overview of the extended models for Europe's performance with XGBoost.

\paragraph{SURR + WTHR + CRY}
For pH the values are similar in $CaCl_2$ and $H_{2}O$, highlighting similar accuracy between both test methods with values of 0.86 and 0.81 and thus being 22\% better than in BASE configuration. Also Phosphorus (24.88), Nitrogen (3.40), and Potassium (200.42) show a better RMSE than the BASE configuration with 7\% improvement. It can also be mentioned, that all RMSE values are slightly to moderately better than the standard deviation.

\paragraph{SURR + WTHR + CRY + CLAY}
The values of RMSE for pH (0.90 and 0.85) are 18\% lower than compared to the BASE model. Phosphorus (25.05), Nitrogen (3.44), and Potassium (202.03) are 6\% lower compared to the BASE model. Additionally, also with this configuration all RMSE values are slightly to moderately better than the corresponding standard deviation.

\paragraph{Effects of CLAY}
When comparing \textit{SURR + WTHR + CRY} against \textit{SURR + WTHR + CRY + CLAY}, it is clear that \textit{SURR + WTHR + CRY} performs slightly better than \textit{SURR + WTHR + CRY + CLAY} across all nutrients. Hence for the XGBoost approach Clay may attribute negative effects to the models' performances due to the large number of extra features (1024) compared to other features all together.\\

\begin{table}[htbp]
\caption{Extended Model for Europe: XGBoost Performance}
\begin{center}
\begin{tabular}{|c@{}|c@{}|c@{}|@{}c@{}|@{}c@{}|@{}c@{}|}
\hline
\multicolumn{1}{|l|}{\multirow{5}{*}{\textbf{Nutrient}}} & \multicolumn{1}{l|}{\multirow{5}{*}{\textbf{Unit}}} & \multicolumn{1}{l|}{\multirow{5}{*}{\textbf{\begin{tabular}[c]{@{}l@{}}Mean \\ ± StdDev\end{tabular}}}} & \multicolumn{3}{@{}c@{}|}{\textbf{RMSE}}                                                                                                                                                                                                                                                           \\ \cline{4-6} 
\multicolumn{1}{|l|}{}                                   & \multicolumn{1}{l|}{}                               & \multicolumn{1}{l|}{}                                                                                   & \multicolumn{1}{c|}{\multirow{4}{*}{\textbf{BASE}}} & \multicolumn{1}{c@{}|}{\multirow{4}{*}{\textbf{\begin{tabular}[c]{@{}c@{}}Previous+ \\ SURR, \\ WTHR, \\ CRY\end{tabular}}}} & \multicolumn{1}{l|}{\multirow{4}{*}{\textbf{\begin{tabular}[c]{@{}l@{}}Previous+ \\ CLAY\end{tabular}}}} \\
\multicolumn{1}{|l|}{}                                   & \multicolumn{1}{l|}{}                               & \multicolumn{1}{l|}{}                                                                                   & \multicolumn{1}{c|}{}                               & \multicolumn{1}{c|}{}                                                                                                      & \multicolumn{1}{l|}{}                                                                                     \\
\multicolumn{1}{|l|}{}                                   & \multicolumn{1}{l|}{}                               & \multicolumn{1}{l|}{}                                                                                   & \multicolumn{1}{c|}{}                               & \multicolumn{1}{c|}{}                                                                                                      & \multicolumn{1}{l|}{}                                                                                     \\
\multicolumn{1}{|l|}{}                                   & \multicolumn{1}{l|}{}                               & \multicolumn{1}{l|}{}                                                                                   & \multicolumn{1}{c|}{}                               & \multicolumn{1}{c@{}|}{}                                                                                                      & \multicolumn{1}{l|}{}                                                                                     \\ \hline
pH in CaCl2 & - & 5.71 ± 1.40 & 1.09 & \textbf{0.86} & 0.90 \\
pH in H2O & - & 6.26 ± 1.32 & 1.03 & \textbf{0.81} & 0.85 \\
Phosphorus & mg/kg & 26.95 ± 27.02 & 26.53 & \textbf{24.88} & 25.05 \\
Nitrogen & g/kg & 3.15 ± 3.70 & 3.63 & \textbf{3.40} & 3.44 \\
Potassium & mg/kg & 204.83 ± 208.25 & 216.48 & \textbf{200.42} & 202.03 \\
\hline
\end{tabular}
\label{tab_extended_performance_xgboost}
\end{center}
\end{table}

\subsubsection{Fully Connected Neural Networks}
Table \ref{tab_extended_performance_fcnn} shows an overview of the Extended Model for Europe's performance with FCNNs.

\paragraph{SURR + WTHR + CRY}
The pH values in $CaCl_2$ and $H_{2}O$ are very similar with values of 0.93 and 0.87 resulting in 17\% and 23\% better performance compared to the BASE configuration. Also Phosphorus (24.37), Nitrogen (3.27), and Potassium (159.03) achieve a better RMSE than the BASE configuration with 11\%, 5\%, and 15\% improvement. All RMSE values are slightly to moderate better than the standard deviation for each nutrient.

\paragraph{SURR + WTHR + CRY + CLAY}
The values for pH (0.91 and 0.90) are 19\% lower in comparison to the BASE configuration. For Phosphorus the RMSE (23.06) is 15\% lower compared to BASE, whereas Nitrogen's RMSE (3.35) is only reduced by 3\%. Potassium's RMSE (166.36) is reduced by 11\% and all RMSE values are again slightly to moderate better than the standard deviation.

\paragraph{Effects of CLAY}
We compare \textit{SURR + WTHR + CRY} against \textit{SURR + WTHR + CRY + CLAY} and the results are not as clear as previously with XGBoost. For the FCNN architecture \textit{SURR + WTHR + CRY} as well as \textit{SURR + WTHR + CRY + CLAY} both reach top RMSE values. \textit{SURR + WTHR + CRY} performs better for pH in $H_20$, Nitrogen, and Potassium, whereas \textit{SURR + WTHR + CRY + CLAY} performs better for pH in $CaCl_2$, and Phosphorus. Hence for the FCNN, Clay cannot be attributed a constant positive or negative effect to the models' performances.\\

\begin{table}[htbp]
\caption{Extended Model for Europe: FCNN Performance}
\begin{center}
\begin{tabular}{|c@{}|c@{}|c@{}|c@{}|c@{}|c@{}|}
\hline
\multicolumn{1}{|l|}{\multirow{5}{*}{\textbf{Nutrient}}} & \multicolumn{1}{l|}{\multirow{5}{*}{\textbf{Unit}}} & \multicolumn{1}{l|}{\multirow{5}{*}{\textbf{\begin{tabular}[c]{@{}l@{}}Mean \\ ± StdDev\end{tabular}}}} & \multicolumn{3}{c|}{\textbf{RMSE}}                                                                                                                                                                                                                                                           \\ \cline{4-6} 
\multicolumn{1}{|l|}{}                                   & \multicolumn{1}{l|}{}                               & \multicolumn{1}{l|}{}                                                                                   & \multicolumn{1}{c@{}|}{\multirow{4}{*}{\textbf{BASE}}} & \multicolumn{1}{c@{}|}{\multirow{4}{*}{\textbf{\begin{tabular}[c]{@{}c@{}}Previous+ \\ SURR, \\ WTHR, \\ CRY\end{tabular}}}} & \multicolumn{1}{l|}{\multirow{4}{*}{\textbf{\begin{tabular}[c]{@{}l@{}}Previous+ \\ CLAY\end{tabular}}}} \\
\multicolumn{1}{|l|}{}                                   & \multicolumn{1}{l|}{}                               & \multicolumn{1}{l|}{}                                                                                   & \multicolumn{1}{c|}{}                               & \multicolumn{1}{c|}{}                                                                                                      & \multicolumn{1}{l|}{}                                                                                     \\
\multicolumn{1}{|l|}{}                                   & \multicolumn{1}{l|}{}                               & \multicolumn{1}{l|}{}                                                                                   & \multicolumn{1}{c|}{}                               & \multicolumn{1}{c|}{}                                                                                                      & \multicolumn{1}{l|}{}                                                                                     \\
\multicolumn{1}{|l|}{}                                   & \multicolumn{1}{l|}{}                               & \multicolumn{1}{l|}{}                                                                                   & \multicolumn{1}{c|}{}                               & \multicolumn{1}{c|}{}                                                                                                      & \multicolumn{1}{l|}{}                                                                                     \\ \hline

pH in CaCl2 & - & 5.71 ± 1.40 & 1.12 & 0.93 & \textbf{0.91} \\
pH in H2O & - & 6.26 ± 1.32 & 1.12 & \textbf{0.87} & 0.90 \\
Phosphorus & mg/kg & 26.95 ± 27.02 & 27.12 & 24.37 & \underline{\textbf{23.06}} \\
Nitrogen & g/kg & 3.15 ± 3.70 & 3.44 & \underline{\textbf{3.27}} & 3.35 \\
Potassium & mg/kg & 204.83 ± 208.25 & 185.31 & \underline{\textbf{159.03}} & 166.36 \\
\hline
\end{tabular}
\label{tab_extended_performance_fcnn}
\end{center}
\end{table}

\subsubsection {Random Forest}
Table \ref{tab_extended_performance_rf} shows an overview of the extended models for Europe's performance with Random Forest.

\paragraph{SURR + WTHR + CRY}
The RMSE values for pH in $CaCl_2$ and $H_{2}O$ (0.85 and 0.80) are 22\% better than the BASE configuration. Phosphorus with an RMSE of 24.60 and Nitrogen with an RMSE of 3.37 are 7\% better than BASE. Potassium's RMSE (192.01) is 12\% better compared to BASE. While observing the standard deviation all RMSE values are slightly to moderately better than the standard deviation.

\paragraph{SURR + WTHR + CRY + CLAY}
For pH in $CaCl_2$ and $H_{2}O$ the RMSE (0.89 and 0.84) are 18\% better than the BASE RMSE values. Phosphorus (24.80) and Nitrogen (3.42) have a 6\% better RMSE than BASE. Potassium with an RMSE of 197.36 is improved by 9\% in comparison to BASE.
All RMSE values are slightly to moderate better than the standard deviation.

\paragraph{Effects of CLAY}
A comparison of \textit{SURR + WTHR + CRY} to \textit{SURR + WTHR + CRY + CLAY} shows, that for all nutrients \textit{SURR + WTHR + CRY} performs better than \textit{SURR + WTHR + CRY + CLAY}. Hence for the Random Forest approach Clay can be attributed with a slightly negative effect to the model's performance.

\begin{table}[htbp]
\caption{Extended Model for Europe: Random Forest Performance}
\begin{center}
\begin{tabular}{|c@{}|c@{}|c@{}|c@{}|c@{}|c@{}|}
\hline
\multicolumn{1}{|l|}{\multirow{5}{*}{\textbf{Nutrient}}} & \multicolumn{1}{l|}{\multirow{5}{*}{\textbf{Unit}}} & \multicolumn{1}{l|}{\multirow{5}{*}{\textbf{\begin{tabular}[c]{@{}l@{}}Mean \\ ± StdDev\end{tabular}}}} & \multicolumn{3}{c|}{\textbf{RMSE}}                                                                                                                                                                                                                                                           \\ \cline{4-6} 
\multicolumn{1}{|l|}{}                                   & \multicolumn{1}{l|}{}                               & \multicolumn{1}{l|}{}                                                                                   & \multicolumn{1}{c@{}|}{\multirow{4}{*}{\textbf{BASE}}} & \multicolumn{1}{c@{}|}{\multirow{4}{*}{\textbf{\begin{tabular}[c]{@{}c@{}}Previous+ \\ SURR, \\ WTHR, \\ CRY\end{tabular}}}} & \multicolumn{1}{l|}{\multirow{4}{*}{\textbf{\begin{tabular}[c]{@{}l@{}}Previous+ \\ CLAY\end{tabular}}}} \\
\multicolumn{1}{|l|}{}                                   & \multicolumn{1}{l|}{}                               & \multicolumn{1}{l|}{}                                                                                   & \multicolumn{1}{c|}{}                               & \multicolumn{1}{c|}{}                                                                                                      & \multicolumn{1}{l|}{}                                                                                     \\
\multicolumn{1}{|l|}{}                                   & \multicolumn{1}{l|}{}                               & \multicolumn{1}{l|}{}                                                                                   & \multicolumn{1}{c|}{}                               & \multicolumn{1}{c|}{}                                                                                                      & \multicolumn{1}{l|}{}                                                                                     \\
\multicolumn{1}{|l|}{}                                   & \multicolumn{1}{l|}{}                               & \multicolumn{1}{l|}{}                                                                                   & \multicolumn{1}{c|}{}                               & \multicolumn{1}{c|}{}                                                                                                      & \multicolumn{1}{l|}{}                                                                                     \\ \hline
pH in CaCl2 & - & 5.71 ± 1.40 & 1.09 & \underline{\textbf{0.85}} & 0.89 \\
pH in H2O & - & 6.26 ± 1.32 & 1.02 & \underline{\textbf{0.80}} & 0.84 \\
Phosphorus & mg/kg & 26.95 ± 27.02 & 26.50 & \textbf{24.60} & 24.80 \\
Nitrogen & g/kg & 3.15 ± 3.70 & 3.63 & \textbf{3.37} & 3.42 \\
Potassium & mg/kg & 204.83 ± 208.25 & 216.06 & \textbf{192.01} & 197.36 \\
\hline
\end{tabular}
\label{tab_extended_performance_rf}
\end{center}
\end{table}

\subsubsection{Model Performance Comparison}
We compare all three models to each other to identify the best performing model and configuration for each nutrient. \\
XGBoost with both \textit{SURR + WTHR + CRY} as well as \textit{SURR + WTHR + CRY + CLAY} configuration performs less across all nutrients when compared to FCNN and Random Forest. \\
The FCNN approach outperforms the Random Forest approach for Phosphorus, Nitrogen, and Potassium. Interestingly for Phosphorus \textit{SURR + WTHR + CRY + CLAY} performs better than \textit{SURR + WTHR + CRY}. This implies that Clay has positive effect to reduce the RMSE at least for Phosphorus. Hence \textit{SURR + WTHR + CRY + CLAY} with the FCNN should be picked for Phosphorus. For both Nitrogen and Potassium the \textit{SURR + WTHR + CRY} configuration performs better without Clay by achieving its best RMSE with the FCNN approach. Hence, for the nutrients Nitrogen and Potassium \textit{SURR + WTHR + CRY} with the FCNN approach should be selected. \\
For pH in $CaCl_2$ and $H_{2}O$ the Random Forest \textit{SURR + WTHR + CRY} without clay performs the best and should be used for prediction.\\

\subsubsection{Feature Importance}

\begin{figure}[ht]
    \centering
    \includegraphics[width=1.0\linewidth]{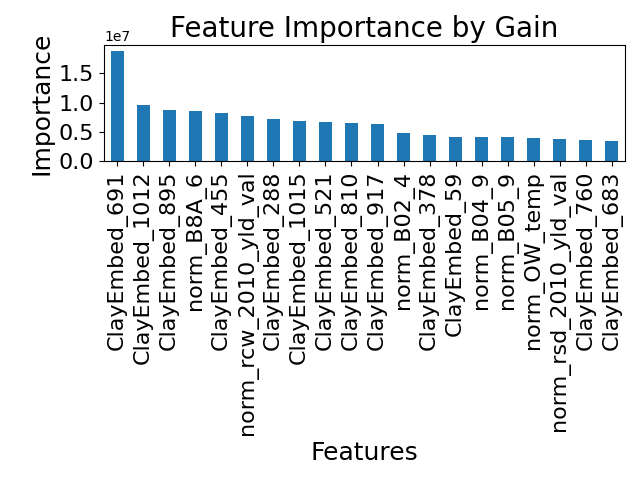}
    \caption{Comparison of the 20 most important features of the extended data sets for K, based on the extended XGBoost model}
    \label{fig:04_FeatImp-K}
\end{figure}

Further, we want to achieve an insight into the influence of the input data on the model performance. The feature importance for the extended model based on the XGBoost is shown in the following figures. Due to the high number of input features, only the top 20 most important features are plotted.\\
In this analysis for potassium (Figure \ref{fig:04_FeatImp-K}), the feature ‘Clay embedding\_691’ is the most important (highest gain), followed by two other Clay embeddings; a Sentinel band appears at rank 4, and a yield feature ranks 6th.

\begin{figure}[ht]
    \centering
    \includegraphics[width=1.0\linewidth]{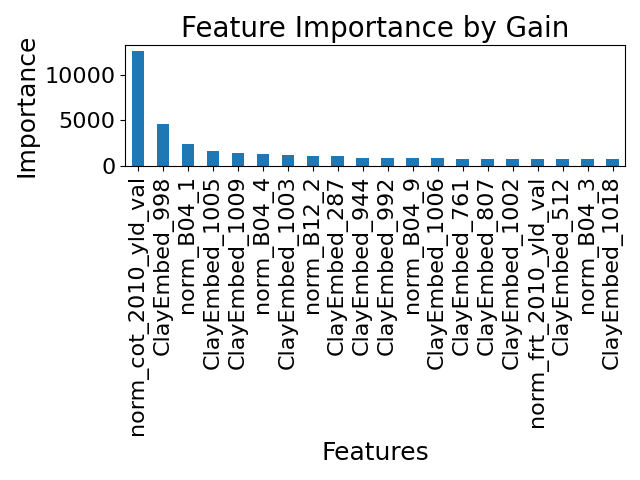}
    \caption{Comparison of the 20 most important features of the extended data set for N, based on the extended XGBoost model}
    \label{fig:04_FeatImp-N}
\end{figure}

Nitrogen as next target value has a Yield input data as the most important feature, namely norm\_cot\_2010\_yld\_val, which is the actual yield for cotton, as can be seen in figure \ref{fig:04_FeatImp-N}. The next important features, with a relatively lower gain, are Clay embeddings and Sentinel bands.

\begin{figure}[ht]
    \centering
    \includegraphics[width=1.0\linewidth]{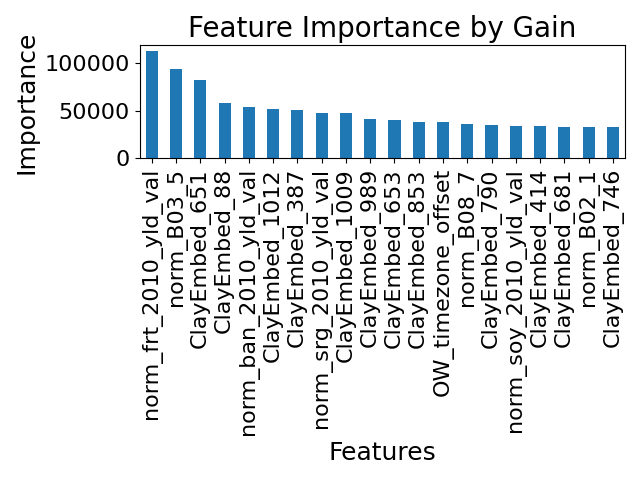}
    \caption{Comparison of the 20 most important features of the extended data set for P, based on the extended XGBoost model}
    \label{fig:04_FeatImp-P}
\end{figure}

The resulting feature importance for phosphorus in figure \ref{fig:04_FeatImp-P} shows features with a very similar gain. Starting with norm\_frt\_2010\_yld\_val (fruits) a yield feature is most important. Followed by a Sentinel band, two Clay embeddings, and an additional yield input.

\begin{figure}[ht]
    \centering
    \includegraphics[width=1.0\linewidth]{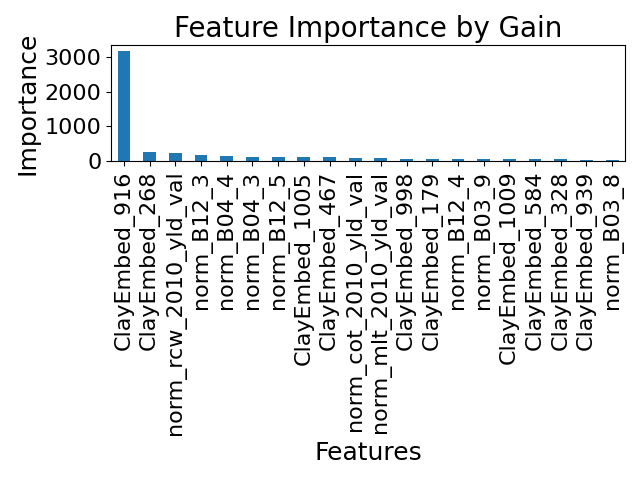}
    \caption{Comparison of the 20 most important features of the extended data set for pH from H2O, based on the extended XGBoost model}
    \label{fig:04_FeatImp-H2O}
\end{figure}

\begin{figure}[ht]
    \centering
    \includegraphics[width=1.0\linewidth]{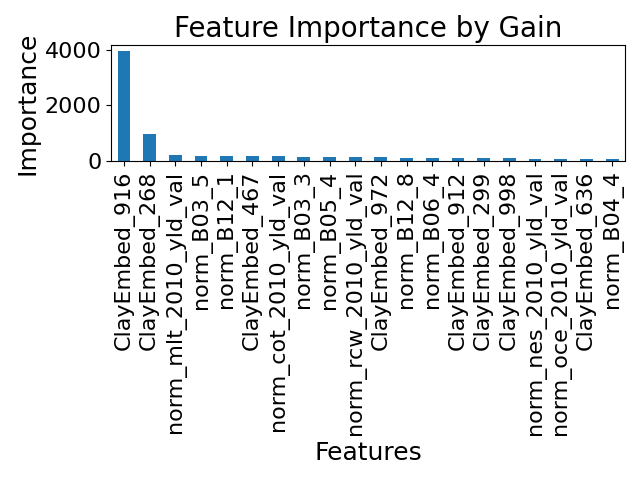}
    \caption{Comparison of the 20 most important features of the extended data set for pH from CaCl2, based on the extended XGBoost model}
    \label{fig:04_FeatImp-CaCl2}
\end{figure}

For the two pH predictions, the feature importance for the \(H_2O\)-based model is presented in Figure \ref{fig:04_FeatImp-H2O} and for the \(CaCl_2\)-based model in Figure \ref{fig:04_FeatImp-CaCl2}. It is observed that the most and second most important features for both pH values are the same clay embeddings. The third-ranked feature is a yield input, although not identical between the two models. The subsequent two features are Sentinel bands in both cases. Overall, the feature importance for both pH predictions is very similar, which aligns with the high correlation between the two pH target values.

In conclusion, it is clear that Clay embeddings have a high impact on the feature importance, even if they don't always improve the prediction quality as discussed in the previous section.\\

\section[B]{Challenges in Extending on Africa}
\label{adapafrica}
As described in Section \ref{ChapterMethTargetData}, a key problem with African soil data is the absence of timestamps, which makes it impossible to map a soil sample to a timely, accurate satellite image.

\subsection[B-Input-Data]{Input Data}
The currently available WoSIS dataset predates the launch of Sentinel-2, making it necessary to search for an alternative satellite imagery source. Only around 200 WoSIS data points outside of Europe fall within the timeframe since the launch of Landsat 8; therefore, we shift our focus to using Landsat 7 images. However, additional challenges come along, as Landsat 7 suffers from a Scan Line Corrector (SLC) failure since 2003, causing gaps in the resulting satellite images \cite{Landsat7}. Obtaining historical data for Landsat 7 was also difficult, as the Copernicus Data Space provided only a limited number of datasets. In order to address this, the existing data acquisition scripts are extended to use the USGS M2M API, the official source of Landsat data \cite{USGS_M2M}. Unfortunately, while the API allows searching and retrieving metadata, its download endpoint is disabled, and no further information is provided regarding the reason. The last remaining free source for Landsat 7 images is Google Earth Engine, but integrating it would have required significant modifications to the existing data collection scripts, which would be beyond the scope of the project \cite{LandsatEE}.
Due to the challenges mentioned above, the following approaches could not be validated. Nonetheless we hope that the following project continues from this point once adequate satellite data become available..

\subsection[B-Target-Data]{Target Data}

The global soil data retrieved from WoSIS is shown in TABLE \ref{tab_soil_data}. WoSIS (World Soil Information Service) \cite{WoSIS2024} is a global initiative led by the International Soil Reference and Information Centre (ISRIC) with the primary purpose of facilitating the harmonization, collection, and dissemination of soil data. It has more than 228,000 worldwide profiles available which have been collected between 1920 and 2020. However, over 74\% of WoSIS profiles were collected at depths greater than 30 cm. For our analysis, only soil profiles with a maximum depth of 25 cm are considered, since satellite imagery can only capture information from the topsoil. Additionally, almost 27\% of WoSIS soil profiles don't have timestamps. After filtering based on these criteria, ensuring Landsat 7/8 data availability (after March 2008), and excluding European data (to avoid overlap and bias with the LUCAS 2018 TOPSOIL dataset), only 3,641 soil profiles remain. Unfortunately, within this subset, only one soil dataset is located in Africa.
 
\begin{table}[htbp]
\caption{WoSIS 2023 snapshot profiles with timestamp and Landsat 7 and 8 coverage by continent}
\begin{center}
\begin{tabular}{| l | c | c | c |}
\hline
\textbf{Continent} & \textbf{pH} & \textbf{P} & \textbf{N} \\
\hline
Africa & 1 & 0 & 1 \\
Asia & 34 & 26 & 34 \\
North America & 2,708 & 806 & 1,615 \\
Oceania & 366 & 2 & 155 \\
South America & 21 & 3 & 21 \\
\hline
\end{tabular}
\label{tab_wosis_data}
\end{center}
\end{table}

\subsection{Model Validation}
Due to the scarcity of African soil data and the lack of timestamps, two approaches are suggested to validate whether our extended model for Europe generalizes for Africa. Both approaches require soil data by continents and therefore we separated the previously processed WoSIS profiles by continent as shown in TABLE \ref{tab_wosis_data}.\\

\subsubsection{Extended Model for Europe with World Data (without Africa)}\label{ref_europe_with_world_data}
The idea for the future project group would be to cross validate the extended model for Europe - which is trained with European data only - with data from other continents around the world, once satellite data becomes available. As part of this comparison, the test RMSE values of European data should be compared to the test RMSE values, when the model is run with data from other continents. Our hypothesis is that if the test RMSE values are similar for each continent, then the extended model for Europe generalizes well to other regions, including Africa.\\

\subsubsection{Extended Model for Europe with African Soil Data}
Another approach for the future project group could be to wait for African soil data that has accurate timestamps available and use them to further train and tune our existing extended models for Europe. One very promising project to solve this missing soil data gap is the \textit{Soils4Africa} project by the European Commission, which runs until 31/05/2025. At the time of this project, no data have been published yet from the Soils4Africa project \cite{Soils4Africa}.\\

\section{Discussion and Summary}

\subsection[Discussion]{Result Discussion}
The research demonstrates that splitting the modeling process by training separate models for each target soil parameter is feasible. This approach results in lower training times, reduced model complexity, and improved prediction accuracy with lower error. In contrast, a single model trained on all target values would face additional missing-value challenges, since not all input data are available for every target, which would ultimately reduce overall performance.

Although spatial data are used as input, our investigation of the train, validation, and test splits shows that the models are not highly sensitive to the splitting method. However, spatial cross-validation is considered the preferred approach, as it better accounts for spatial dependencies.

The evaluated and optimized models exhibit very comparable prediction quality across most nutrients. The only noticeable difference occurs for potassium, where the FCNN outperforms the other models. 

The expansion of the database with additional features clearly improves model quality, as evidenced by the feature importance maps of the extended input data. Therefore, incorporating more data is generally recommended. Nonetheless, an excessive number of input features can increase model complexity and prolong training times. For example, while Clay embeddings multiply the number of input features, only a few of these embeddings appear to be highly influential. Future improvements might include compressing the embeddings to a lower dimension using an autoencoder or eliminating less important features.

Regarding prediction quality, the achieved RMSE values for all target variables (except pH) are in a similar range as the corresponding means. Although these errors may seem high, they can be partly explained by the strong skew in the data toward lower values rather than a normal distribution. Moreover, the fact that the RMSE values are within the range of the standard deviation indicates that the prediction quality is not yet fully satisfactory. However, the error levels are comparable to those reported in \cite{Hengl2021}, suggesting that the model quality is at least on par with existing references—or possibly that the reference errors are themselves high, as noted by \cite{Schut2020}. Despite these challenges, the results justify continued pursuit of this project.

\subsection[Summary]{Project Summary}
In summary, the machine learning methods applied in this project indicate significant potential for predicting soil quality. Numerous types of data can be used as features for prediction, and while there is room for further model improvement, any extension must be carefully managed to keep the input scope reasonable and the training and inference times feasible.

The next step is to determine what range of errors is tolerable for a realistic prediction, and whether the observed errors are low enough to support crucial fertilization recommendations. If not, targeted model improvements and refined data selection will be necessary based on the insights gained so far.

As discussed in Section \ref{adapafrica}, one of the biggest challenges remains the collection of African soil data with accurate timestamps corresponding to the available satellite data. Without resolving this issue, there will always be a quality gap for soil predictions in Africa, where inexpensive and easily accessible fertilizer recommendations are critically needed.

Further recommendations and ideas for improvements are discussed in the subsequent section.

\section[Outlook]{Outlook}

During the research for this project, new ideas and opportunities for further improvement have emerged, which are outlined below.

\subsection{Further Data Improvements}

As is inherent in machine learning, the quality of the data exerts a strong influence on the performance of any prediction system. This principle applies to soil prediction in general and, specifically, to the task presented here. In this context, the following suggestions are made:

\begin{itemize}
    \item One of the most serious challenges for an African prediction model is the absence of soil data with accurate timestamps. This limitation makes it difficult to match soil samples with temporally relevant satellite imagery, and it may account for critical views on such approaches \cite{Schut2020}. Addressing this issue would represent a major improvement over the research in \cite{Hengl2021}. An enhanced database is expected from the Soils4Africa project, which is scheduled for completion in 2025 \cite{Soils4Africa}.
    
    \item The time interval between soil sample collection and the acquisition of the corresponding satellite image—especially for surface soil properties that are highly sensitive to short-term weather effects such as rain—should be minimized to improve prediction quality.
    
    \item The LUCAS soil data, particularly for phosphorus, contain many values below the limit of detection. Employing a random imputation strategy within the range from zero to the detection limit, rather than using a constant value, might further enhance prediction accuracy.
    
    \item The team responsible for the open weather data source used in this study has expressed interest in the research and in pursuing further projects. This opens up the possibility of gaining access to additional or more detailed weather data.
    
    \item Performing spatial cross-validation using clusters determined by, for example, a k-nearest-neighbor algorithm rather than the current grid-based clustering might improve prediction performance by better capturing spatial dependencies.
    
    \item Numerous features have been extracted from satellite data for other applications (e.g., NDVI \cite{NDVI2025}). These features, or the algorithms used to derive them, could potentially be leveraged for soil prediction.
    
    \item Additional and more recent data sources—such as satellite data from future measurement campaigns like LSTM on temperature (Land Surface Temperature Monitoring, scheduled to start in 2029, \cite{LSTM2025}) and CIMR on climate change (Copernicus Imaging Microwave Radiation, scheduled to start in 2028, \cite{CIMR2025}) as recommended in \cite{Hengl2021}—could be incorporated.
    
    \item The amount of sunlight during satellite image acquisition significantly influences result quality. Therefore, an algorithm that filters out satellite images captured under low sunlight conditions would be beneficial. Low sunlight conditions might occur, for example, from September to March in northern European regions or in small valleys.
    
    \item To extend the range of input data, the use of local plant images for plant health estimation—as recommended in \cite{Ennaji2023}—could be explored. For instance, users of a future app could directly upload images of their own plants to the prediction system, providing an additional, highly localized input feature.
    
    \item Additional information provided by users, such as details on crop rotation, previous fertilization practices, and the application of artificial irrigation, may further improve predictions.
    
    \item Crop rotation history might also be inferred from satellite data, as suggested in \cite{Ennaji2023}.
    
    \item The current world yield data could be further refined by accounting for the use of artificial irrigation.
    
    \item Open-source satellite data with higher resolution (up to 30 cm per pixel) are currently lacking; however, such data could lead to additional improvements.
    
    \item Higher-resolution satellite data could also be utilized in the clay model to obtain improved embeddings as input.
    
    \item Embeddings could be generated as a separate pre-processing step using autoencoders \cite{Diederik2019}, particularly for neighboring pixels in satellite data. Additionally, autoencoders could be used to reduce the dimensionality of the clay embeddings, resulting in a more proportional representation of the input data and potentially leading to improved performance and reduced training times.

    \item The FAO GAEZ portal’s “Theme 5: Actual Yields and Production” dataset may be supplemented by data from “Theme 6: Yield and Production Gaps,” which provide estimates of how much additional yield could be achieved under optimal conditions. Initial experiments suggest that these yield gap figures can further improve nutrient predictability. However, it is important to confirm the independence of the Theme 6 dataset from ground-truth observations; if it already incorporates actual soil measurements, it may inflate model performance without providing genuinely new information.
\end{itemize}

\subsection{Possible Machine Learning Enhancements}
The analysis of feature importance presented for the XGBoost model reveals that, due to the nonlinearity and ambiguous behavior inherent in tree-based models, these importance results can vary significantly with different data and training runs. This variability is partially reflected in the plots shown earlier. Consequently, a more sophisticated and less model-biased method for determining feature importance is desirable. The use of permutation feature importance \cite{Altmann2010} is recommended, as it can provide a more robust assessment. With this measure, reducing the number of input features in the extended model becomes conceivable, which would not only speed up training times but also facilitate future predictions on handheld devices.

To realize a soil prediction app, the implementation of an ML-Ops strategy \cite{Akkem2023} is advisable. Although ML-Ops was not implemented in the current project due to the lack of database updates, future stages of app development—especially upon rollout—should include tracking and monitoring of datasets, deployments, and additional processes. Accordingly, the use of MLflow \cite{MLflow2025} as a framework in combination with Python is suggested. As an alternative, albeit not open source, Neptune \cite{Neptune2025} is also recommended.

\subsection{Further Topics}
Beyond the aforementioned ideas for improvements in data and models, broader topics must also be considered for the successful implementation of a fertilization recommendation system. In particular, the method by which information regarding fertilization needs is communicated to farmers is crucial. As stated by \cite{Ennaji2023}, "The real problem is understanding the right path to the best nutrient management recommendations and making them accessible and understandable to farmers."

\section*{Acknowledgments}
The authors would like to thank:
\begin{itemize}
%    \item Prof. Dr. Torsten Schön from Technische Hochschule Ingolstadt for supervising the project and providing valuable ML expertise.
%    \item Dr. Denis Dalic from MI4People for establishing the project idea and delivering the initial information.
    \item Prof. Dr. Patrick Noack from Hochschule Weihenstephan-Triesdorf for his expert knowledge and for engaging in insightful discussions on fertilization and nutritional values.
    \item Maximilian Schmailzl for providing his private server as computational resource for the model trainings
\end{itemize}

\section*{Disclaimer}
Throughout the development of this report, the project team utilized ChatGPT as a supplementary tool to streamline tasks such as technical description drafts and text refinement. Final responsibility for the research design, analysis, and conclusions remains entirely with the authors. All content generated by ChatGPT underwent careful review, validation, and, where necessary, revision to ensure it met academic standards and reflected the team's original work.

\bibliography{bib}
\bibliographystyle{unsrt}

\end{document}